\documentclass[final,5p,twocolumn]{elsarticle}
\biboptions{sort&compress}
\usepackage[utf8]{inputenc}
\usepackage{amsmath}
\usepackage{amssymb}
\usepackage{graphicx}
\usepackage{acro}
\usepackage[hyphens]{url}
\usepackage{hyperref}
\usepackage[capitalize]{cleveref}
\usepackage{enumitem}
\usepackage{multirow}
\usepackage{siunitx}
\usepackage{float}
\usepackage[flushleft]{threeparttable}
\usepackage[table,xcdraw]{xcolor}
\usepackage[switch]{lineno}
\usepackage{longtable}
\usepackage{placeins}
\usepackage{tikz}
\DeclareAcronym{CSF}{
    short = CSF,
    long = cerebrospinal fluid
}

\DeclareAcronym{dHCP}{
    short = dHCP,
    long = Developing Human Connectome Project
}

\DeclareAcronym{MRI}{
    short = MRI,
    long = Magnetic Resonance Image
}

\DeclareAcronym{MR}{
    short = MR,
    long = Magnetic Resonance
}

\DeclareAcronym{CNN}{
    short = CNN,
    long = Convolutional Neural Network
}

\DeclareAcronym{VINN}{
    short = VINN,
    long = Voxel-size Independent Neural Network
}

\DeclareAcronym{VINNA}{
    short = VINNA,
    long = \ac{VINN} with "internal augmentations"
}

\DeclareAcronym{infantFS}{
    short = infantFS,
    long = infantFreeSurfer
}

\DeclareAcronym{iBEAT}{
    short = iBEAT,
    long = Infant Brain Extraction and Analysis Toolbox
}

\DeclareAcronym{FastInfantSurfer}{
    short = FIS,
    long = FastInfantSurfer 
}
\acuse{FastInfantSurfer}

\DeclareAcronym{WM}{
    short = WM,
    long = white matter
}

\DeclareAcronym{GM}{
    short = GM,
    long = gray matter
}    

\DeclareAcronym{PVE}{
    short = PVE,
    long = partial volume effect}
    
\DeclareAcronym{abide-ii}{
    short = ABIDE-II,
    long = Autism Brain Imaging Data Exchange II}
    
\DeclareAcronym{NN}{
    short = NN,
    long = nearest-neighbour
}

\DeclareAcronym{ASD}{
    short = ASD,
    long = Average Surface Distance
}

\DeclareAcronym{DSC}{
    short = DSC,
    long = Dice Similarity Coefficient
}

\DeclareAcronym{TODO}{
    short = TODO,
    long = compile fix still todo
}

\DeclareAcronym{HiRes}{
    short = HiRes,
    long = high-resolution
}

\DeclareAcronym{LowRes}{
    short = LowRes,
    long = low-resolution
}

\DeclareAcronym{MSDA}{
    short = MSDA,
    long = multi-source domain adaptation
}

\setcitestyle{square}

\journal{Imaging Neuroscience}

\newcommand{\PyTorch}{\mbox{PyTorch}}

\newcommand{\new}[1]{#1}
\newcommand{\newfigure}[1]{#1}

\begin{document}
\begin{frontmatter}

\title{VINNA for Neonates - Orientation Independence through Latent Augmentations }

\author[label1]{Leonie Henschel}
\author[label1]{David Kügler}
\author[label2,label3]{Lilla Zöllei\corref{cor1}}
\author[label1,label2,label3]{Martin Reuter\corref{cor1}\corref{cor2}}
\cortext[cor1]{Equal Contribution.}
\cortext[cor2]{Corresponding author.}
\address[label1]{German Center for Neurodegenerative Diseases (DZNE), Bonn, Germany}
\address[label2]{A.A.\ Martinos Center for Biomedical Imaging, Massachusetts General Hospital, Boston MA, USA }
\address[label3]{Department of Radiology, Harvard Medical School, Boston MA,USA}


\begin{abstract}
A robust, fast, and accurate segmentation of neonatal brain images is highly desired to better understand and detect changes during development and disease, specifically considering the rise in imaging studies for this cohort. Yet, the limited availability of ground truth datasets, lack of standardized acquisition protocols, and wide variations of head positioning in the scanner pose challenges for method development. A few automated image analysis pipelines exist for newborn brain MRI segmentation, but they often rely on time-consuming non-linear spatial registration procedures and require resampling to a common resolution, subject to loss of information due to interpolation and down-sampling. Without registration and image resampling, variations with respect to head positions and voxel resolutions have to be addressed differently.
In deep-learning, external augmentations such as rotation, translation, and scaling are traditionally used to artificially expand the representation of spatial variability, which subsequently increases both the training dataset size and robustness. However, these transformations in the image space still require resampling, reducing accuracy specifically in the context of label interpolation. 
We recently introduced the concept of resolution-independence with the Voxel-size Independent Neural Network framework, VINN. Here, we extend this concept by additionally shifting all rigid-transforms into the network architecture with a four degree of freedom (4-DOF) transform module, enabling resolution-aware internal augmentations (VINNA) for deep learning. In this work we show that VINNA (i)~significantly outperforms state-of-the-art external augmentation approaches, (ii)~effectively addresses the head variations present specifically in newborn datasets, and (iii)~retains high segmentation accuracy across a range of resolutions (0.5-\SI{1.0}{\milli\meter}). Furthermore, the 4-DOF transform module together with internal augmentations is a powerful, general approach to implement spatial augmentation without requiring image or label interpolation. The specific network application to newborns will be made publicly available as VINNA4neonates.


%


\end{abstract}
\begin{keyword}
Computational Neuroimaging \sep Deep Learning \sep Structural MRI \sep Artificial Intelligence \sep High-Resolution \sep Newborn Brain
\end{keyword}
\end{frontmatter}

\section{Introduction}

Collections of neonatal brain \acp{MRI} are indispensable to understand brain development and to detect early signs of potential developmental disorders. One of the key tasks in \ac{MRI} analysis is automated segmentation, the labeling of anatomical regions of interest (ROIs) that can be used for quantitative modeling of healthy development, for analyses in population studies, for understanding disease effects as well as a starting point for further neuroimaging tasks. The segmentation of infant \acp{MRI} is a challenging and non-trivial undertaking due to the rapid non-linear changes during the postnatal brain growth period, elevated levels of head motion, limited availability of congruent datasets, varying intensity profiles across scanners, protocols and modalities, as well as the inversion of gray-white contrast around the age of 5-9 month \cite{ajayi-obe_reduced_2000,prastawa_automatic_2005,dubois_early_2014,gilmore_longitudinal_2012}. In this paper, we focus on a sub-group of the infant population -- newborns -- and present a four-degree of freedom (4-DOF) transform module to address two core challenges within this cohort: non-uniform image resolutions (scaling) and increased variability of head positions (rigid transformations = rotation and translation) during image acquisition. The 4-DOF transform module is directly integrated into the network-architecture and addresses the variability of head positions internally. As such, it expands and generalizes the distinguishing feature, resolution independence, of the recently published \ac{VINN} \cite{Henschel_2022} by rotation and translation transformations. We refer to our new framework as \ac{VINNA}.   

\begin{figure*}[!hbt]
    \centering
    \includegraphics[width=\textwidth,keepaspectratio]{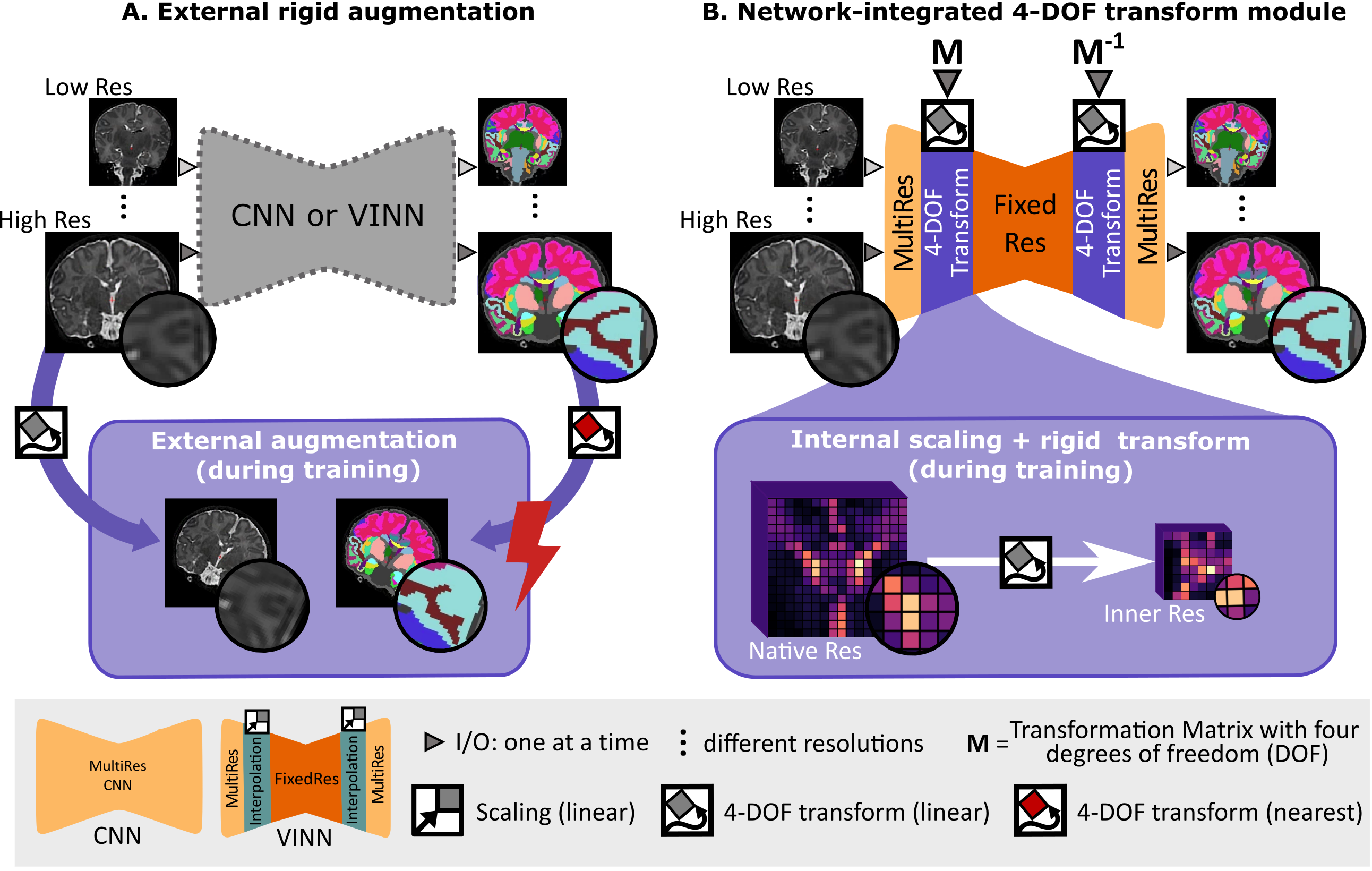}
    \caption{Spatial augmentations in deep learning networks: \textbf{A.}~One single resolution-ignorant CNN or resolution-independent voxel size independent network (VINN) can learn to segment multiple resolutions and head positions by training on a diverse dataset. External scale, rotation, and translation augmentation (+ external augmentation, A.\ bottom) diversifies the existing training samples by resampling the images and the reference segmentation maps. Here, however, lossy interpolation and resulting artefacts, especially from nearest-neighbour interpolation of discrete label maps, may result in a loss of structural details and sub-optimal performance. \textbf{B.}~Our 4-DOF transform module (VINNA) completely avoids interpolation of the images and discrete labels by integrating the interpolation step into the network architecture itself (B.\ bottom). Rotations, translations, and scalings applied in the first interpolation block are later reversed, assuring segmentations to be in the original image orientation. Furthermore, the explicit transition from the native resolution to a normalized internal resolution facilitates an understanding of the difference between image features (MultiRes blocks with distances measured in voxels) and anatomical features (FixedRes inner blocks with normalized distances). }
    \label{fig:networks}
\end{figure*}

In contrast to adults, newborn head positions in the scanner are far more diverse due to the scanning conditions (asleep) and overall smaller anatomy. 
While padding is often used to stabilize the child's head and to occupy the space between head coil and participant (e.g.\ foam cushions, pads, mats, pillows, or visco-elastic matters) \cite{copeland_infant_2021}, its standardization is difficult. This results in diverse head orientations within the scanner and potentially high variations among imaging studies.

In addition, there is no de-facto standard resolution for newborn imaging. In the case of research protocols, when more time is available, \acp{MRI} are often acquired at higher-resolutions to address the small size of brain structures and stronger partial volume effects \cite{dubois_2019,dubois_2021,cbf_pc_mri_res2019}.
However, the range of recorded resolutions across research and clinical studies is relatively large and heterogeneous, ranging from \SI{0.5}{\milli\meter} to \SI{3.0}{\milli\meter} in past and recent research studies (e.g.\ NIH-PD \cite{EVANS2006184}, BCP \cite{bcp}, \ac{dHCP} \cite{dhcp_pipeline2018,dhcp_diffusion}, HBCD \cite{hbcd}). Similarly, resolutions are not standardized across atlases (UNC 0-1-2 Infant atlases \cite{unc2011}, Imperial Brain Development atlases \cite{imperial1_2011,imperial2_2012,gousias-2012,imperial3_2016}), that are often used to guide the automated labeling algorithms. 

Traditional tools for newborn or infant segmentation predominantly interpolate images to a single chosen resolution and harmonize the spatial head position via atlas registrations \cite{ibeat,ourselin_automatic_2015,prieto_multiseg_2019,infantfs,dhcp_pipeline2018,shi_construction_2010}. Resampling of images can, however, result in loss of information, especially in the context of high-resolution label maps. Furthermore, atlas-guided approaches are usually highly dependent on registration accuracy. For newborns, registration is particularly challenging due to lower tissue contrast, specifically in the T1w scans. Errors in the process are hence common and improvement of the registration, e.g.\ with spatio-temporal information, anatomical constraints, and surface models, is an active field of research \cite{makropoulos-2014,shi_construction_2010,imperial1_2011,dubois_2021,ahmad2019,lebenberg_framework_2018,garcia2018,bozek2018}. 

The explicit definition of spatial and intensity features can be avoided by using \acp{CNN}. In fact, fast deep-learning methods for semantic segmentation, are becoming increasingly popular for infant segmentation \cite{wang_ibeat_2023,dolz_deep_2020,qamar_variant_2020,nie_fully_2016,zhang_deep_2015,zeng_2023,Zeng2018,kumar_infinet_2018,wang_multi-task_2019,moeskops_automatic_2015,wang_id-seg_2022}. Applicability of deep learning approaches, however, is generally restricted to domains where sufficiently large training datasets exist. While there have been several initiatives to collect larger neuroimaging cohorts of newborns and infants in recent years \cite{dhcp_pipeline2018,bcp,hbcd,dhcp_diffusion}, their size is still relatively small compared to equivalent cohorts in the adult population. Additionally, accompanying manual labels are sparse, due to high annotation costs (time and money) and non-uniform labeling protocols, limiting the pool for supervised training options further. Considering the newborn cohort, the variability in resolution and head positioning is likely underrepresented in the publicly available datasets, questioning whether a network trained on the available pairs of scans and labels can be robust enough without additional augmentation.

The most widely used solution to artificially increase the training set size, robustness, and generalizability of deep-learning methods has been traditional data augmentation, such as rotation, scaling, or translation (\Cref{fig:networks}A). In this case, both images and their labelmaps are interpolated to a new random position during training. Interpolation, however, in the native image space requires resampling of the discrete ground truth segmentations, resulting in information loss (e.g.\ from lossy \ac{NN} interpolation) and reduction in accuracy \cite{Henschel_2022}. 

With the \ac{VINN} architecture \cite{Henschel_2022}, we recently established the first network for resolution-independent deep learning, which effectively circumvents scaling augmentation and subsequent external resampling, while leveraging information across datasets of varying resolutions. 
In \ac{VINN}, the classic fixed-factor integer down- and up-scale transitions, often implemented via pooling operations in UNet-like architectures \cite{Ronneberger2015}, are replaced with a flexible re-scaling for the first and last scale transitions. This network-integrated resolution-normalization allows for segmentation in the native space during both, training and inference. In adults, this approach has been shown to outperform fixed-resolution \acp{CNN} as well as resolution-ignorant \acp{CNN} trained with external scaling augmentation, and to improve performance both for sub-millimeter and one-millimeter scans. 

Since newborn datasets are often acquired at various native resolutions and are particularly subject to partial volume effects, the resolution-normalization feature offers a basis to improve segmentation performance here as well. 
As is, \ac{VINN} only addresses scaling and would still require external augmentations, and hence label interpolation, to address the increased variability of head positions and limited availability of training data for newborns.
With our \ac{VINNA} and its 4-DOF transform module, we now close this gap and propose a paradigm shift away from classical data augmentation towards a detail-preserving internal augmentation scheme (\Cref{fig:networks}B). While avoiding any type of label interpolation, we extend \ac{VINN}´s network-integrated resolution-normalization with spatial augmentations (i.e., rotation and translation). At the first layer scale transition, the feature maps are hence not only flexibly rescaled, but also randomly transformed to imitate the position variations commonly found in newborns, subsequently increasing the training distribution.  


In conclusion, \new{\ac{VINNA} and its 4-DOF transform module effectively address} the challenges associated with newborn segmentation, namely variation in head positions and resolutions in the context of limited data availability. 
\new{The four key contributions of \ac{VINNA} presented in this work are as follows:}
\new{\begin{enumerate}[label=(\roman*)]
     \item We provide the first publicly available open-source deep-learning pipeline for a combined subcortical segmentation as well as cortical, and white matter parcellation for newborn T1w or T2w \acp{MRI}.
    \item We introduce a novel augmentation paradigm, which for the first time moves spatial augmentations into the network (instead of being performed outside). Our experimental results compare various spatial augmentation approaches side-by-side to isolate their effects.
    \item We ensure fair comparisons throughout, for example, by fixed dataset splits, retraining methods under equal data and parameter settings, comparing architectures and setups with minimal differences, and quantifying real-world performance
    \item We, further, provide extensive comparison with state-of-the-art deep-learning methods (2D and 3D nnUNet) adapted for newborn segmentation (retrained on the same data) and
    present an extensive comparison to the publicly available newborn segmentation pipelines iBEAT and infantFS.
\end{enumerate}
}
The specific application of \ac{VINNA} to newborns will be made available as VINNA4neonates within our open source repository\footnote{\url{github.com/Deep-MI/NeonateVINNA} upon publication} including Docker containers offering easy accessibility for the community.


\subsection{Related work}
While various reliable and sensitive traditional \cite{FSL, FSLFast, spm_book2007,fischl2002whole} and  fast deep-learning solutions exist \cite{wachinger2018deepnat,quicknat,slant2019,Mehta2017,Chen2018VoxResNet,Sun2019,Ito2019,MeshNet,AssemblyNet2020,Billot_2020,Iglesias_2021,Henschel_2020,Henschel_2022} for adult whole brain segmentation, application of these methods to younger ages is hampered by the significant differences in size, \ac{MRI} contrast profiles and rapidly changing postnatal anatomy that is challenging to model with static templates.

\subsection{Traditional tools for infant segmentation}
Infant-specific traditional atlas-guided tools \cite{shi_construction_2010,ourselin_automatic_2015,prieto_multiseg_2019,infantfs,dhcp_pipeline2018,beare_neonatal_2016} are predominantly optimized for a specific age range, resolution, and modality. Further, they differ significantly in the number of segmented classes and structure definitions.

The more recent \ac{iBEAT} V2.0 \cite{wang_ibeat_2023} is a combination of age-specific \acp{CNN} for tissue segmentation, traditional surface generation and parcellation,  based on atlas registration, into 34 regions following the Desikan-Killiany protocol \cite{Desikan2006}. It supports a large age range (0-6 years), and allows segmentation of both, T1w and T2w \ac{MRI}. While multiple input resolutions are supported, \ac{iBEAT} internally reorients and resamples each image to a standard format (RAS orientation and \SI{0.8}{\milli\meter} isotropic resolution). Hence, it does not support native resolution segmentation and image interpolation is required to map segmentations back to the original input space. The resampling step is automatically included in the pipeline such that in- and output resolutions are flexible. Furthermore, in its publicly available docker pipeline\footnote{\url{https://github.com/iBEAT-V2/iBEAT-V2.0-Docker}}, segmentation is limited to \ac{WM}, \ac{GM} and \ac{CSF}. 


\ac{infantFS} \cite{infantfs}, on the other hand, mimics the FreeSurfer \cite{fischl2012freesurfer} processing pipeline for adults and processes images from the first two years postnatally. It supports anatomical segmentation into 32 classes based on multi-atlas label fusion strategy including registration to the infantFreeSurfer training data set \cite{atlas_infantfs}. The entire pipeline is publicly available\footnote{\url{https://surfer.nmr.mgh.harvard.edu/fswiki/infantFS}} and allows processing of T1w images at a resolution of \SI{1.0}{\milli\meter}, where the atlas training data is defined. For newborns, T1w images often suffer from poor tissue contrast due to the underlying myelination process, aggravating accurate registration from the atlases onto individual brains. This age group can therefore be a challenge for \ac{infantFS}´s mono-modality approach.

The \ac{dHCP} minimal-processing-pipeline \cite{dhcp_pipeline2018} is an optimized framework for cortical and sub-cortical volume segmentation, cortical surface extraction, and cortical surface inflation, which has been specifically designed for high-resolution T2w \acp{MRI} of newborns \cite{hughes_dedicated_2017}. Here, an Expectation-Maximization algorithm, including an atlas-based spatial prior term, labels 87 classes based on a modified version of the ALBERTs atlas \cite{gousias-2012,makropoulos-2014}. The segmentations include subcortical structures, cortical and \ac{WM} parcellations. Due to the cubic increase in voxel-size for high-resolution images, processing times are in the order of hours to days for a single subject. This is a common limitation among traditional methods. 

\subsection{Deep-Learning for infant segmentation}
\subsubsection{Newborns}
Overall, networks for cortical parcellations and subcortical structure segmentations in newborns are limited. The few existing \acp{CNN} support a single modality (T2w), fixed resolution, and segment a single \cite{rajchl2017} or eight \cite{moeskops_automatic_2015} tissue classes. One recent exception is the deep-learning based neuroimaging pipeline by Shen et al.\ \cite{shen_automatic_2023}, which is trained with the \ac{dHCP} data. Here, the authors propose a 3D multi-task deep learning model with a U-Net like architecture to segment structural T1w and T2w images on both thin and thick sliced images. Unfortunately, the network follows a fixed-resolution scheme (\SI{0.8}{\milli\meter}), it does not support native segmentation across resolutions commonly encountered in newborn cohorts, and it is not readily available online. 

\subsubsection{Isointense phase}
The vast majority of deep-learning tools focus on processing of images at the isointense phase around 6 months after birth \cite{dolz_deep_2020,qamar_variant_2020,nie_fully_2016,zhang_deep_2015,zeng_2023,Zeng2018,kumar_infinet_2018,9328318,sadegh_2020}. Via the iSeg-challenge \cite{wang_benchmark_2019, sun_multi-site_2021}, data for training and validation is conveniently available  partly explaining this predominance. While many interesting architectural solutions have arisen, the main focus of the works is the effective combination of information from both T1w and T2w images to generate a broad segmentation into \ac{CSF}, \ac{GM} and \ac{WM}. \new{This modality combination is specifically important in the isointense phase, which is characterized by changes in the myelination strongly effecting the appearance of the recorded \acp{MRI} \cite{wang2015,weisenfeld2009automatic,gui2012morphology}. The inversion of the \ac{WM}-\ac{GM} signal results in extremely low tissue contrast. The newborn cohort, on the other hand, demonstrates good contrast between \ac{GM} and \ac{WM}, specifically on the T2w images. While the age difference is small, the two cohorts as well as the associated challenges are distinct and networks trained on the one can not easily be applied to the other. Subsequently, neither resolution-independence nor the stronger variation of head positions are specifically accounted for in network solutions for the isointense phase.}

\subsubsection{Cross-age generalizability}
To address generalizability across different age groups, recent research has suggested optimized training strategies for neonates, such as multi-task learning of tissue segmentation and geodesic distances \cite{wang_multi-task_2019} or the use of inductive biases in the form of pre-trained weights (i.e.\ fine-tuning to the target domain) \cite{wang_id-seg_2022}. Both approaches improve segmentation accuracy, but they are still limited in their generalizability. They require retraining and hence a sufficient amount of labeled data; additionally they rely on private datasets, limiting their reproducibility. Recently, a contrast agnostic segmentation via synthetic images, originally proposed for adult brain segmentation \cite{Billot_2020,Iglesias_2021}, has also been adopted for infant segmentation \cite{shang_learning_2022}. Unfortunately, the output resolution is fixed for the network, and native resolution segmentations are not supported. Furthermore, while the model was able to generalize across a broader age range, the synthetic images still differ considerably from real data and the network therefore underperformed compared to age-specific models trained on existing \acp{MRI}.
 
\subsection{Resolution-independence and position transforms in deep-learning}

A general resolution-ignorant framework addressing position transforms via external augmentations is nnUNet \cite{isensee_2021}. This network has successfully been applied for a variety of segmentation tasks due to its inherent ability to construct optimal parameter settings based on the input data itself. It provides different network set-ups (2D, 3D and a cascaded 3D approach for large images) as well as a number of external image augmentations including random rotation, scaling, mirroring, and gamma transformation. Interestingly, while the trained network also follows a fixed-resolution scheme, pre- and post-processing automatically resamples between original image and network resolution. While native resolution segmentation is not supported, in- and output resolutions are not fixed and the method is therefore a valid alternative to our \ac{VINNA}. Both, the 2D and 3D nnUNet + exA therefore serve as a state-of-the-art baseline for the newborn segmentation task. 

\new{A siamese network for semi-supervised training of a network to become equivariant to elastic transformation has been proposed in a single-resolution setting \cite{Bortsova2019SemisupervisedMI}. A dedicated loss function assures that segmentations are consistent under a given class of transformation applied first to the image, and second to the output. The approach therefore relies on external augmentation and applies the transformation in the image space (before and after a UNet). The proposed VINNA, on the other hand, is fundamentally different. It shifts this step into the network itself, hence creating an internal augmentation. Overall, the approach by Bortsova et al \cite{Bortsova2019SemisupervisedMI} does therefore assure consistency across transformations in the labeling space, while VINNA targets spatial consistency of the feature maps.}

In spatial transformers \cite{spatialT}, transformations attempt to harmonize or re-orient the image into a better position. To this end, an affine transformation is implicitly learned via a dedicated localisation network. Subsequent application of the calculated coordinate grid resamples the source feature maps via bi-linear interpolation to the new position. While our approach shares grid calculation and interpolation within the network with spatial transformers, our internal augmentation approach is inherently different. First, spatial transformers do not diversify or augment feature maps, but rather try to reach a harmonized position with respect to the data seen during training. External augmentations are still necessary to expose the network to a wide data variety and approximate equivariance. Second, instead of a localisation network, we directly determine the sampling-grid based on a transformation matrix, which allows for an explicit integration of knowledge about the image, such as the resolution. As a result, computational complexity is reduced while achieving the desired position diversification and resolution-independence.

\section{Material and methods}
\begin{figure}[!h]
    \centering
    \newfigure{\includegraphics[width=\columnwidth, keepaspectratio]{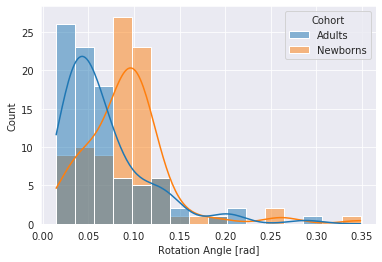}}
    \caption{Comparison of variation in head position in newborns (orange) and adults (blue). Newborns show greater variation with respect to rotation angles. Rotation transformation is based on alignment of each individual subject to their midspace. N=90 for both cohorts}
    \label{fig:rot-cohort}
\end{figure}

\begin{figure*}[!hbt]
    \centering
    \includegraphics[width=\textwidth,keepaspectratio]{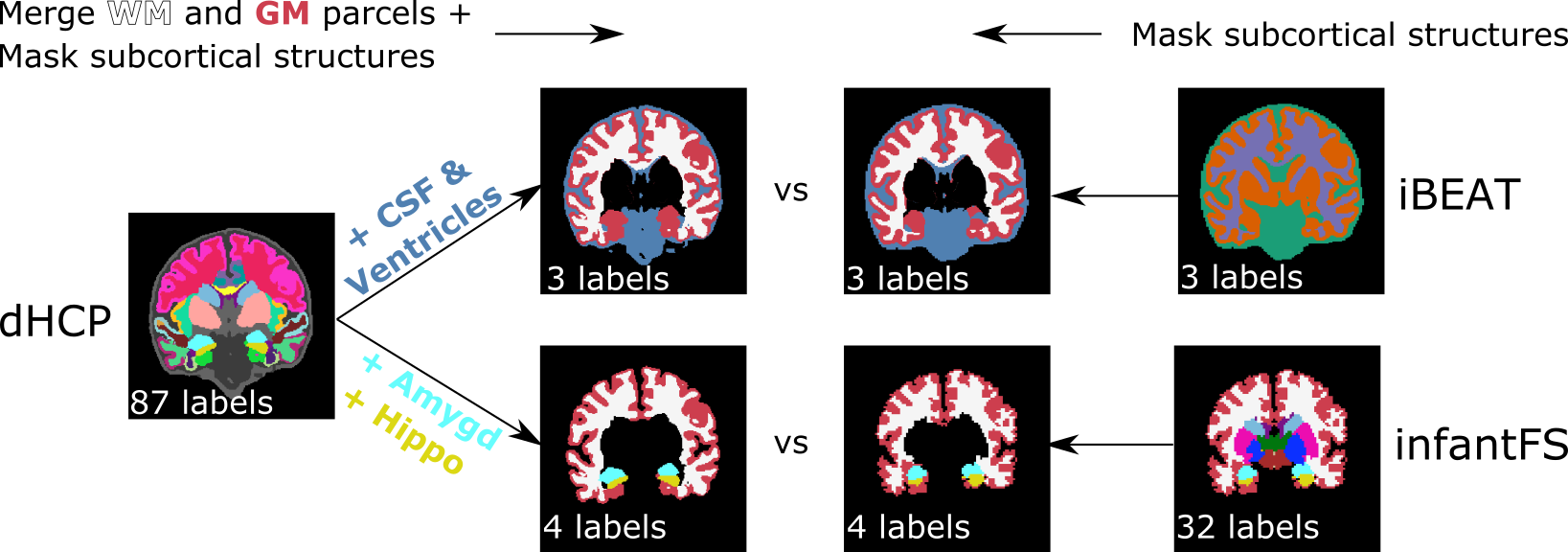}
    \caption{Harmonization of inconsistent label protocols between \ac{iBEAT}, \ac{infantFS}, and \ac{dHCP}. Reduction of the original \ac{dHCP} segmentation from 88 labels (left) by first merging all cortical parcels to cortex, \ac{WM} parcels to \ac{WM} and removal of subcortical structures, is followed by addition of \ac{CSF} and ventricles (top) or amygdala and hippocampus (bottom) for comparison to \ac{iBEAT} (3 labels: \ac{CSF} (blue), \ac{WM} (white) and \ac{GM} (red), top second from left top) or \ac{infantFS} (4 labels: \ac{WM} (white), cortex (red), hippocampus (yellow) and amygdala (blue), bottom second from left), respectively. For \ac{iBEAT} (right top) all \ac{GM} and \ac{WM} are modified using the \ac{dHCP} segmentation for the subcortical structures resulting in three labels for the final mapped version (3 labels, top second from right). For \ac{infantFS} (right bottom), all structures except \ac{WM}, cortex, hippocampus, and amygdala are removed (4 labels, bottom second from right).}
    \label{fig:labelmap}
\end{figure*}

\subsection{Datasets}\label{sec:datasets}
As the largest publicly available newborn dataset with intensity images, accompanying subcortical segmentations as well as cortical and \ac{WM} parcellations at the time of our experiments, we \new{randomly assign} participants from the \ac{dHCP} cohort with corresponding T1w and T2w \acp{MRI} \new{to the} training, testing, and validation \new{sets, while ensuring equal distribution of age and gender}. Additionally, the Melbourne Children’s Regional Infant Brain (M-CRIB) atlas cohort serves as an independent testing set for external validation of the final method. Written informed consent to participate in this study was provided by the participants’ legal guardian or next of kin in accordance with the Institutional Review Board. Complete ethic statements are available at the respective study webpages.

\vspace{1ex}
    \textbf{dHCP}: The developing Human Connectome Project ~\cite{dhcp_pipeline2018} includes T1w and T2w \acp{MRI} of newborns imaged without sedation on a 3 T Philips Achieva scanner. It provides \SI{0.5}{\milli\meter} isotropic de-faced scans of individuals imaged postnatally between 24 to 45 weeks post-conceptional age. Imaging data for 578 participants with matching T2w and T1w were selected. The original images were acquired in sagittal and axial slice stacks with in-plane resolution of \SI{0.8}{\milli\meter} $\times$ \SI{0.8}{\milli\meter} and \SI{1.6}{\milli\meter} slices overlapped by \SI{0.8}{\milli\meter}. Motion correction and super-resolution reconstruction techniques \cite{cordero-grande-2018,Kuklisova-Murgasova-2012} created isotropic volumes of resolution \SI{0.5}{\milli\meter}. All T1w scans follow the same inversion recovery multi-slice fast spin-echo protocol with TR \SI{4.795}{\second}, TE \SI{8.7}{\milli\second}, TI \SI{1.740}{\second}, and SENSE factor 2.27 (axial) / 2.56 (sagittal). The parameters for the T2w scans are TR \SI{12}{\second}, TE \SI{156}{\milli\second} and SENSE factor 2.11 (axial) / 2.66 (sagittal). The full dataset is available online\footnote{\url{https://data.developingconnectome.org/}}. In the present study, 318 quality checked images are used for network training and 90 for validation. A total of 170 images are used in the final test set. 

     \new{Even though the \ac{dHCP} follows a well-defined protocol, standardization of positioning during scanning is still a challenge. As shown in \Cref{fig:rot-cohort}, inter-subject head position diversity in the newborn cohort is larger than an equally standardized adult cohort (HCP).}

\vspace{1ex}
    \textbf{M-CRIB}: The Melbourne Children’s Regional Infant Brain (M-CRIB) atlas \cite{mcrib_2019} is constructed from 10 T2w \ac{MRI} and corresponding manual segmentations of healthy term-born neonates (four females, six males) with gestational age-at-scan between 40.29–43.00 weeks. The atlas comprises 94 neonatal brain regions compatible with the widely-used Desikan-Killiany-Tourville adult cortical atlas \cite{Klein2012}. The T2w \acp{MRI} scanning protocols include the usage of a transverse T2 restore turbo spin echo sequence with \SI{1.0}{\milli\meter} axial slices, a TR of \SI{8.910}{\second}, TE of \SI{152}{\milli\second}, flip angle of 120 degrees, Field Of View of \SI{192}{\milli\meter} × \SI{192}{\milli\meter}, and in-plane resolution of \SI{1}{\milli\meter} (zero-filled interpolated to \SI{0.5}{\milli\meter} × \SI{0.5}{\milli\meter} × \SI{1}{\milli\meter}). The T2w images are bias-corrected, skull-stripped and resampled to \SI{0.63}{\milli\meter} × \SI{0.63}{\milli\meter} × \SI{0.63}{\milli\meter} isotropic voxels. All 10 participants are used as an independent testing set for our external validation experiments.

\subsection{Generation of reference segmentation with the dhcp-minimal-processing-pipeline}\label{sec:ground_truth}
\vspace{1ex}
To imitate various resolutions and create the desired reference segmentations for training, we processed all raw \ac{dHCP} \acp{MRI} with the dhcp-minimal-processing-pipeline \cite{dhcp_pipeline2018} at \SI{1.0}{\milli\meter}, \SI{0.8}{\milli\meter} and \SI{0.5}{\milli\meter}. The structure definitions follow the ALBERTs atlas \cite{gousias-2012} with the subdivision of the \ac{WM} and cortex proposed by Makropolus et al.\ \cite{makropoulos-2014}, resulting in a total of 87 structures (3 background labels, 20 subcortical regions, 32 cortical parcels, 32 \ac{WM} parcels). We further lateralized the intracranial background based on the average Euclidean distance to neighbouring labels resulting in a final count of 88 labels. We provide a list of all segmentation labels used for training in the Appendix (see \Cref{tab:labels}). As the \ac{dHCP} cohort includes both, T2w and a co-registered T1w \acp{MRI}, we trained dedicated networks for each modality. Note, that the dhcp-minimal-processing-pipeline relies on the original T2w images to create its segmentations, which are generally of higher quality in this collection. 

\subsection{Traditional infant segmentation tools}\label{sec:traditional-tools}

To evaluate \ac{VINNA} against state-of-the-art traditional segmentation methods, we further process the testing set with the docker version of the \ac{iBEAT} V2.0 pipeline \cite{wang_ibeat_2023} and \ac{infantFS} \cite{infantfs}. 

\subsubsection{\ac{iBEAT}}
The \ac{iBEAT} V2.0 pipeline \cite{wang_ibeat_2023} combines both traditional and deep-learning models to create tissue segmentations into three classes (\ac{GM}, \ac{WM}, and \ac{CSF}), surface models, and cortical parcellations of the pial surface into 34 regions based on the Desikan-Killiany protocol \cite{Desikan2006}. For tissue segmentation, \ac{iBEAT} uses seven age-specific \acp{CNN} trained on data for the representative age group ($\leq 1$ 
month, 3 months, 6 months, 9 months, 12 months, 18 months, and 24+ months of age). Neither the source code nor the training data and labels are publicly available. Hence, retraining of the models is not possible and comparisons are limited to the \ac{iBEAT} pipeline output as is. For processing with \ac{iBEAT}, submissions via a webserver\footnote{\url{https://ibeat.wildapricot.org/}} or processing with a docker image\footnote{\url{https://github.com/iBEAT-V2/iBEAT-V2.0-Docker}} are possible. The docker version does not currently support the cortical parcellations of the surface models. Due to the large number of participants, privacy concerns, and longer processing times when submitting via the webserver, we decided to use the docker version to process the T2w images of the testing set at the original \SI{0.5}{\milli\meter} resolution. The resulting 3-label tissue segmentations form the basis for comparison to the other tools in this paper.

\subsubsection{\ac{infantFS}}
To allow comparison of segmentation performance to \ac{VINNA}, all available T1w images from the \ac{dHCP} testing set are processed with \ac{infantFS} with default settings. The neuroimaging pipeline \ac{infantFS} creates surface models, anatomical segmentations and cortical parcellations based on the Desikan-Killiany-Tourville atlas \cite{Klein2012} for 0-2 year old infants akin to the version for adults (FreeSurfer \cite{fischl2012freesurfer}). The tool runs on T1w \acp{MRI} at a resolution of \SI{1.0}{\milli\meter} (non-conforming images are resampled). \ac{infantFS} relies on a registration-based  multi-atlas label fusion strategy and returns an anatomical segmentation into 32 classes, including two labels for \ac{GM} and \ac{WM}.  

\subsubsection{Label harmonization}\label{sec:label-harm}
As the \ac{dHCP}-ALBERTs atlas differs from the resulting segmentations of both \ac{iBEAT} and \ac{infantFS}, we merge, remove, and mask classes to reach an approximate consensus across predictions. \Cref{fig:labelmap} shows the merging protocol on a representative participant with the original and mapped ground truth \ac{dHCP} labels (left side) together with \ac{iBEAT} (top right side) and \ac{infantFS} (bottom right side). First, the 32 cortical and \ac{WM} parcels from the \ac{dHCP} ground truth segmentation are reduced to two labels (cortex and \ac{WM}) (top left in \Cref{fig:labelmap}). For \ac{iBEAT}, the \ac{WM} additionally includes the corpus callosum while \ac{GM} also encompasses the hippocampus and amygdala. The \ac{CSF} label corresponds to the union of lateral-ventricles and \ac{CSF} in the \ac{dHCP}-ALBERTs atlas. In the \ac{dHCP} ground truth, these labels are consequently merged to create the final three classes (\ac{GM}, \ac{WM}, and \ac{CSF}; top second to left image). All other subcortical structures without a single possible assignment to \ac{GM}, \ac{WM}, or \ac{CSF} are masked in the \ac{iBEAT} prediction using the \ac{dHCP} ground truth (\Cref{fig:labelmap}, top right two images). For \ac{infantFS}, the hippocampus and amygdala label remain, while individual cortex and \ac{WM} parcels of the \ac{dHCP} ground truth are merged (\Cref{fig:labelmap}, bottom left images). Hence, in the \ac{infantFS} predictions the following four labels remain: cortex, \ac{WM}, hippocampus and amygdala (\Cref{fig:labelmap}, bottom two images to the right).

\subsection{Network architectures}\label{sec:fastsurfer}

\begin{figure*}[!hbt]
    \centering
    \newfigure{\includegraphics[width=\textwidth,keepaspectratio]{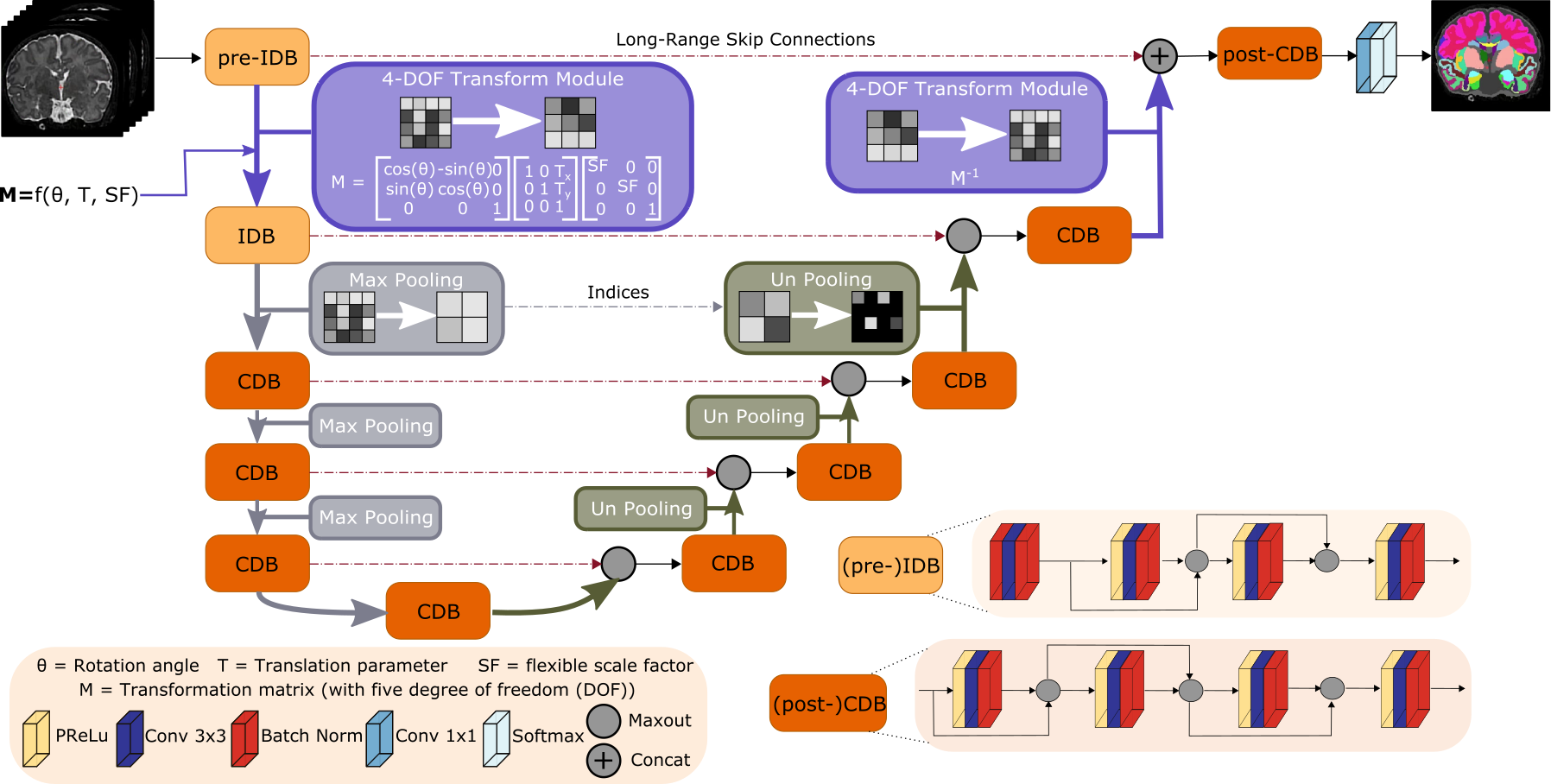}}
    \caption{Network-integrated position variation and scaling normalization in \ac{VINNA}. Flexible transitions between resolutions and head positions become possible by replacing (un)pooling with our network-integrated 4-DOF transform module (purple) after the first input dense block in the encoder (pre-IDB) and before the last competitive dense block in the decoder (post-CDB). A transformation matrix composed of rotation angle $\theta$, translation parameter T and the scaling factor SF defines the feature alterations. Scale transitions between the other competitive dense blocks (CDB) remain standard MaxPool and UnPool operations. Each CDB is composed of four sequences of parametric rectified linear unit (PReLU), convolution (Conv), and batch normalization (BN). In the first two encoder blocks ((pre)-IDB), the PReLU is replaced with a BN to normalize the inputs.}
    \label{fig:multiresnet}
\end{figure*}

\subsubsection{Macro architecture}
\Cref{fig:multiresnet} shows the macro architecture for \ac{VINNA}. While the proposed 4-DOF transform module (purple) can, in theory, be included in any UNet-like architecture, we use the same basic setup for all trained models to assure maximum comparability (i.e.\ same number of parameters, same kernel sizes, etc.). \ac{CNN}*, \ac{VINN}, and \ac{VINNA}, all contain an encoder and decoder consisting of five competitive dense blocks, respectively, which are separated by a bottleneck layer. In the encoder, max pooling operations rescale the feature maps at each level by one half between the blocks. In contrast, index-unpooling doubles the feature map size in the decoder. Skip connections between the blocks at each level allow the gradient to flow efficiently. 
In \ac{CNN}* \cite{Henschel_2022}, pooling and unpooling operations transition between all levels (i.e.\ the purple block in \Cref{fig:multiresnet} is substituted with the gray maxpooling/unpooling operation). 
In \ac{VINN} \cite{Henschel_2022}, the first layer pooling and unpooling operation is replaced with a resolution-normalization module. This network-integrated flexible interpolation step allows transitions between resolutions without restrictions to pre-defined fixed voxel sizes, both during training and inference. Hence, images can be processed at their native resolution without prior resampling. Similar to spatial transformers \cite{spatialT}, the interpolation-based transition is divided into two parts: (i)~calculation of the sampling coordinates (\textit{grid generator}) and (ii)~interpolation operation (\textit{sampler}) to retrieve the spatially transferred output feature maps. Here, the sampling calculation relies only on the scale factor $SF$: the quotient of the resolution information of the inner normalized scale $\text{Res}_\text{inner}$, a tune-able hyperparameter set to \SI{0.8}{\milli\meter} throughout our experiments, and the input image $\text{Res}_\text{native}$. The addition of parameter $\alpha$ sampled from a Gaussian distribution with parameters sigma=0.1 and mean=0 slightly augments the scale factor SF ($SF = \text{Res}_\text{inner}/\text{Res}_\text{native} + \alpha$), introduces small resolution variations to the sampling, and increases the robustness of the latent space interpolation. Specifically, the presence of alpha allows for augmentations at the actual anatomical rather than the voxel size as the normalization of the resolution inside \ac{VINN} disentangles perceived voxel versus actual structure size differences.

\subsubsection{Network-integrated 4-DOF transform module}\label{sec:rotation-variation}
In \ac{VINNA}, the transition is implemented via the new network-integrated 4-DOF transform module shown in purple in \Cref{fig:multiresnet}. Here, the sampling coordinate calculation is based on a transformation matrix $M \in \mathbf{R}^{3\times3}$ with four degrees of freedom, encoding not only scaling, but also in-plane rotation, and translation. Parameters for the rotation angle $\theta \in \mathbb{R}$ and translation $T \in \mathbb{R}^2$ are randomly sampled on the fly during training. The scale factor $SF$ is calculated as in \ac{VINN}s resolution-normalization module, i.e.\ by dividing the inner normalized scale by the image resolution with augmentation by the parameter $\alpha$.  In the first transition step (\Cref{fig:multiresnet}, pre-IDB to IDB), the 
affine per-channel mapping ${M\colon {\mathcal {U}}\to {\mathcal {V}}}$ samples the input feature maps $U \in \mathbb{R}^{H_\text{native}\times W_\text{native}}$ to the output feature maps $V \in \mathbb{R}^{H_\text{inner}\times W_\text{inner}}$. In the final transition step (\Cref{fig:multiresnet}, competitive dense block (CDB) to post-CDB), this spatial transformation is reversed by using the inverse transformation matrix $M^{-1}\colon {\mathcal {V}\to {\mathcal {U}}}$. The interpolation itself is performed by applying a sampling kernel to the input map $U$ to retrieve the value at a particular pixel in the output map $V$. The sampling is identical for each channel, hence, conserving spatial consistency. 

\subsubsection{Network blocks}
\textbf{Competitive Dense Block (CDB) design}
In \ac{VINNA}, a CDB is formed by repetitions of the basic composite function consisting of a probabilistic rectified linear unit (pReLU) activation function, a 3$\times$3 convolution, and a batch-normalization (BN). Feature competition within the block is achieved by using maxout \cite{maxout} instead of concatenations \cite{densenet} in the local skip connections. The maxout operation requires normalized inputs and is therefore always performed after the BN (see position of maxout in CDB design in \Cref{fig:multiresnet}). 

\new{\textbf{Input Competitive Dense Block (IDB) design}
In contrast to the described CDB, the first two network blocks follow a different order of operation. Here, the raw inputs are normalized by first passing them through a BN-Conv-BN combination before adhering to the original composite function scheme (Conv-BN-pReLU) (see \Cref{fig:multiresnet}, IDB).}

\new{\textbf{Pre-IDB} 
The \new{first} encoder block in \ac{VINNA} \new{called pre-IDB} (see \Cref{fig:multiresnet}) transfers image intensity information from the native image to the latent space and encodes voxel size and subject-space-dependent information before the internal interpolation step. The composite function scheme is identical to the IDB and the added prefix simply allows differentiation of the block placements.}

\textbf{Post-CDB}
Akin to the pre-IDB, an additional CDB block in the decoder merges the non-interpolated feature information returned from the pre-IDB skip connection and the upsampled feature maps from the network-integrated 4-DOF transform modules. A concatenation operation combines both feature maps, before passing them to a standard CDB block (see \Cref{fig:multiresnet}, (post-)CDB). After the final 1$\times$1 convolution a softmax operation returns the desired class probabilities. 


\subsection{Loss function}
The network is trained with a weighted composite loss function of logistic loss and Dice loss \cite{sdnet} combined with the high-resolution specific weighting from \ac{VINN} \cite{Henschel_2022}. In short, erosion and dilation of the cortex labels creates a binary mask of the outer cortex, small \ac{WM} strands and deep sulci. Wrong predictions in these areas result in a higher loss, hence guiding the network to focus on areas particularly affected by \ac{PVE}. \new{The original publications' ablation experiments evaluated the impact of the different function elements: the logistic loss and Dice loss combination improves overall segmentation performance  \cite{sdnet}, while the high-resolution weighting leads to higher segmentation accuracy on the cortical parcels \cite{Henschel_2022}.}

\new{If we consider $p_{l,i}(x)$ as the estimated probability of pixel $i$ that belongs to class $l$, $y$ as the corresponding ground truth probability, and $\omega_i$ as the associated weight given to the pixel $i$ based the loss function can be formulated as}

\vspace{-2ex}
\new{\begin{equation}
  \mathcal{L} = {\underbrace{-\sum_{l,i} \omega_i y_{l,i} \log p_{l,i}(x)}_\text{Logistic loss}-\underbrace{\sum_{l\vphantom,} \frac{2\sum_i p_{l,i}(x)y_{l,i}}{\sum_i p_{l,i}(x) + \sum_i y_{l,i}}}_\text{Soft Dice loss}}
\label{eq:4a}
\end{equation}
with $\omega_i = \omega_\text{median freq.} + \omega_\text{gradient} + \omega_\text{GM} + \omega_\text{WM/Sulci}$.}

\new{Here, $\omega_\text{median freq.}$ represents median frequency balancing addressing the class imbalance and $\omega_\text{gradient}$ boundary refinement through a 2D gradient vector \cite{sdnet}, while $\omega_\text{GM}$ and $\omega_\text{WM/Sulci}$ assign higher weights to \ac{PVE} affected areas \cite{Henschel_2022}.
}
\subsection{View aggregation}\label{sec:view_agg}
In order to account for the inherent 3D geometry of the brain, we adopt the 2.5D view aggregation scheme from \cite{Henschel_2020,Henschel_2022} \new{for \ac{CNN}*, VINN, and VINNA}. In short, we train one network instance per anatomical plane and calculate a weighted average of the resulting softmax probability maps. The weight of the sagittal predictions is reduced by one half compared to the other two views to account for the missing lateralization in the sagittal view. In this plane, the network predicts 23 instead of 88 classes.  

\subsection{Augmentations}\label{sec:scale-aug}
\textbf{External augmentation (exA)}
The current state-of-the-art approach to introduce robustness to position changes into neural networks is extensive external augmentation (see \Cref{fig:networks}B). Therefore, we contrast our proposed network-integrated 4-DOF transform module against this approach. We use random transforms with rotation parameters sampled from a uniform distribution of the predefined range -180$^\circ$ to 180$^\circ$ and translation by 0 to 15px to augment images during the training phase and interpolate linearly. For \ac{CNN}* \new{and nnUNet} augmentation also includes sampling of scaling parameters from a uniform distribution of the predefined range 0.8 to 1.15. \ac{VINN}´s resolution-normalization module makes this step obsolete. Every minibatch hence consist of a potentially transformed \ac{MRI} (using bi-linear interpolation) and a corresponding label map (using \ac{NN} sampling). By exposing a network to a large variety of possible image positions during training, orientation-robustness can be achieved. All external augmentation routines are implemented using torchIO \cite{torchio}.  

\textbf{Image intensity augmentation}
To allow generalization outside of the \ac{dHCP} cohort, we apply a number of intensity or texture augmentations on the fly to the training batch, namely bias field changes, random gamma alterations, ghosting, spiking, blurring, and Gaussian noise. Each batch sampled from the original data is transformed by any of the operations above with a probability of 0.4. As before, all augmentations are implemented using torchIO. 

\subsection{Evaluation metrics}
We use the \ac{DSC} \cite{dice_measures_1945,sorensen_method_1948} and \ac{ASD} to compare different network architectures and modifications against each other, and to estimate similarity of the predictions with a number of previously unseen scans. Both are standard metrics to evaluate segmentation performance. We establish improvements by statistical testing (Wilcoxon signed-rank test \cite{wilcoxon} after Benjamini-Hochberg correction \cite{Benjamini_Hochberg} for multiple testing) referred to as "corrected p" throughout the paper.

\subsection{Training setup}
\textbf{Training dataset: } 
For training, we select 318 representative participants from the \ac{dHCP} cohort. Resolutions are equally represented with 106 \acp{MRI} at \SI{1.0}{\milli\meter}, \SI{0.8}{\milli\meter}, and \SI{0.5}{\milli\meter}, respectively. Empty slices are filtered from the volumes, leaving on average 137 single view planes per subject and a total training size of at least 20k images per network. We train all directly compared networks (\new{\ac{CNN}*, \ac{VINN}, \ac{VINNA}, 2D nnUNet and 3D nnUNet}) under the same conditions. 

\textbf{Training parameters: } We implement and train independent models to convergence for the coronal, axial, and sagittal planes with \PyTorch~\cite{Paszke2017}, using one NVIDIA V100 GPU with 32 GB RAM. During training, the modified Adam optimizer \cite{adamW} is used with a learning rate of 0.001. Using a cosine annealing schedule \cite{cosine} with warm restarts, the learning rate is adapted after initially 10 epochs. The epoch offset is subsequently increased by a factor of two. The momentum parameter is fixed at 0.95 to compensate for the relatively small mini batch size of 16 images. To ensure a fair comparison, all networks \new{(\ac{CNN}*, \ac{VINN}, \ac{VINNA}, 2D nnUNet and 3D nnUNet)} have been trained under equal hardware and hyper-parameter settings.


\section{Results}

We group the presentation of results into three blocks: 1.~ablative architecture improvements to determine the best performing module for orientation and position transformation (\Cref{sec:results-rotation}), 2.~performance analysis to comprehensively characterize the advantages of \ac{VINNA} with respect to state-of-the-art traditional atlas- and deep-learning-based methods (\Cref{sec:results-sota}) and 3.\ external validation on M-CRIB \cite{mcrib_2019} to asses generalizability and performance with respect to manual labels (\Cref{sec:results-generalization}).
Following best practice in data-science, we utilize completely separate datasets during the evaluations: the validation set for \Cref{sec:results-rotation} (\Cref{tab:datasets}:~Validation), and various test sets for \Cref{sec:results-sota,sec:results-generalization} (\Cref{tab:datasets}:~Testing). This avoids data leakage and ensures that training, method design decisions, and final testing do not influence each other, which could otherwise lead to overly optimistic results (overfitting).

\subsection{External augmentation vs network-integrated 4-DOF transform module in \ac{VINNA}}\label{sec:results-rotation}
As high variances with respect to head orientations and spatial resolutions are common in newborns and are likely to be underrepresented in the limited available data cohorts, we first compare multiple approaches for extension of the training distribution for accurate (sub)millimeter newborn whole brain segmentation. Traditionally, external data augmentation (exA), such as scaling, rotation, or translation, addresses this problem by interpolating both, the image and label maps, to a new, random position during training. Due to the discrete nature of the label maps, lossy NN interpolation cannot be avoided. In contrast, our new 4-DOF transform module in \ac{VINNA} internally emulates possible head transformations and acts directly on the encoded feature maps. To evaluate effectiveness of the exA versus \ac{VINNA}, we compare \ac{VINNA} with parameter-identical \ac{CNN}*, \ac{VINN}, and \ac{VINNA} equipped with exA. Each subsequent improvement in segmentation performance is confirmed by statistical testing (corrected $p < 0.05$).

\begin{figure*}[!tb]
    \centering
    \includegraphics[width=\textwidth,keepaspectratio]{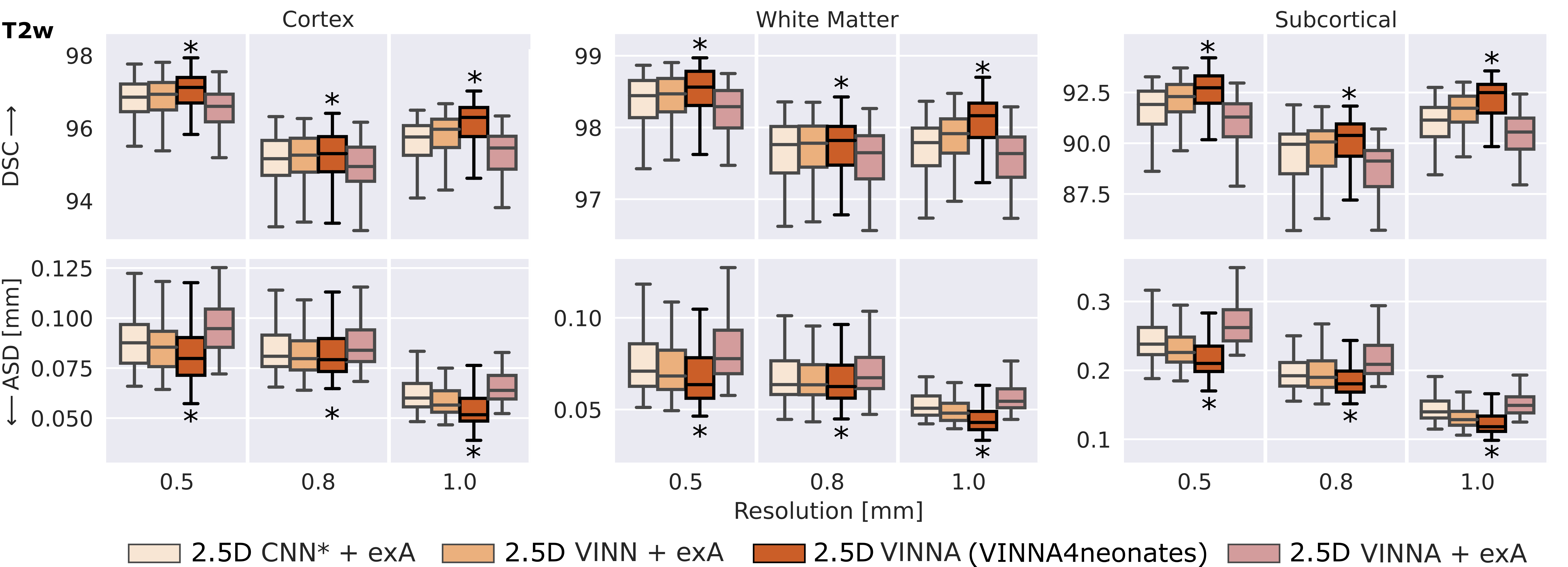}
    \caption{Comparison of approaches for external and internal spatial augmentation on T2w images: Our VINNA method -- with the 4-DOF transform module -- (third box in group) outperforms both state-of-the-art approaches with external Augmentation (CNN* + exA and VINN + exA, first and second box) in Dice Similarity Coefficient (DSC, upper row) and average surface distance (ASD, lower row). This performance advantage is significant (corrected $p < 10^{-9}$, indicated with *) and consistent across resolutions and structures. Combining VINNA with external augmentation (VINNA + exA, last box) reduces performance. 
    exA: external Augmentation.} 
    \label{fig:ablation}
\end{figure*}

In \Cref{fig:ablation}, we compare the model performance of four approaches: \ac{CNN}* and \ac{VINN} with exA (\Cref{sec:fastsurfer}, \ac{CNN}* + exA, left box; VINN + exA, second box from the left), \ac{VINNA} with the new 4-DOF transform module (\ac{VINNA}, second box from the right) and finally \ac{VINNA} with exA (\ac{VINNA} + exA, right box). The analysis of the \ac{DSC} (top) and \ac{ASD} (bottom) is grouped for three groups of structures (cortex averages 32 labels, \ac{WM} averages 32 labels, and subcortical structures average 20 labels) and three resolutions (from left to right \SI{0.5}{\milli\meter}, \SI{0.8}{\milli\meter} and \SI{1.0}{\milli\meter}). We present performance for T2w \acp{MRI}, but we found the same ranking for T1w \acp{MRI}\footnote{To reduce redundancy, we relegate that analysis to the Appendix (\Cref{sec:appendix_ablation_t1}).}.

Looking at the T2w segmentation and focusing on the different resolutions, the differences between the approaches are largest on the subcortical structures. The \ac{CNN}* with exA reaches an average \ac{DSC} of 91.91, 89.95 and 91.14 and an \ac{ASD} of 
\SI{0.238}{\milli\meter}, \SI{0.192}{\milli\meter} and \SI{0.193}{\milli\meter} for input data of \SI{0.5}{\milli\meter}, \SI{0.8}{\milli\meter} and \SI{1.0}{\milli\meter}, respectively. \new{The slight reduction in performance for \SI{0.8}{\milli\meter} resolution consistently occurs for all evaluated models and is probably caused by the necessary image resampling from the original resolution of \SI{0.5}{\milli\meter} and subsequent reprocessing with the dhcp-pipeline (\Cref{sec:ground_truth}). Interpolation from \SI{0.5}{\milli\meter} to \SI{1.0}{\milli\meter} results in a well aligned grid due to the even division by factor 2 (8 voxels get averaged into a single larger voxel). Resampling to \SI{0.8}{\milli\meter} on the other hand, requires an uneven interpolation grid with weighted averages and original voxels that contribute to multiple larger voxels. This more challenging setting could result in the slightly reduced segmentation performance.} Optimization of the architecture design towards multi-resolution (VINN, \Cref{sec:fastsurfer}) leads to significant improvement in the \ac{DSC} and \ac{ASD} across the cortical, \ac{WM}, and subcortical structures (\Cref{fig:ablation}, \ac{VINN} + exA). Particularly, the subcortical segmentations are improved by around 0.5\%. Importantly, the internal 4-DOF transform module (\ac{VINNA}), which avoids label interpolation all together, further reduces the error by one half and increases segmentation performance significantly compared to both \ac{CNN}* and \ac{VINN} with exA. This effect is consistent across all structures and resolutions. Specifically, the \ac{ASD} at the high-resolution benefits from the new module. Here, performance can be improved by 4.47\% on the cortex, 5.19\% on the \ac{WM} and 5.89\% on the subcortical structures. For the lower resolution, the improvement on the cortex and \ac{WM} is slightly lower (around 2\%) while the subcortical structures benefit from the 4-DOF transform module similarly to the \SI{0.5}{\milli\meter} resolution experiments. Overall, \ac{VINNA} reaches the highest \ac{DSC} and lowest \ac{ASD} for the cortical structures (96.24, \SI{0.079}{\milli\meter}), \ac{WM} (98.18, \SI{0.063}{\milli\meter}) and subcortical structures (91.87, \SI{0.190}{\milli\meter}) across all resolutions. The addition of exA to the framework  (\ac{VINNA} + exA, right box in each plot) again reduces performance. The \ac{DSC} drops by 0.4~\%, 0.28~\% and 1.15~\% on the cortex, \ac{WM} and subcortical structures on average, while the \ac{ASD} worsens by 4.75~\%, 3.92~\% and 4.84~\%. Overall, results with \ac{VINNA} are significantly better compared to all ablations on the validation set (corrected $p < 10^{-9}$). 		


						
\subsection{Comparison to state-of-the-art neonate segmentation tools}\label{sec:results-sota}

To evaluate how \ac{VINNA} compares to state-of-the-art neonate MRI segmentation tools, namely  nnUNet (3D and 2D), \ac{iBEAT}, and \ac{infantFS}, we take a closer look at the \ac{DSC} and \ac{ASD} on the testing sets.  

\subsubsection{Comparison of deep-learning networks}
\begin{figure*}[!tb]
    \centering
    \includegraphics[width=\textwidth,keepaspectratio]{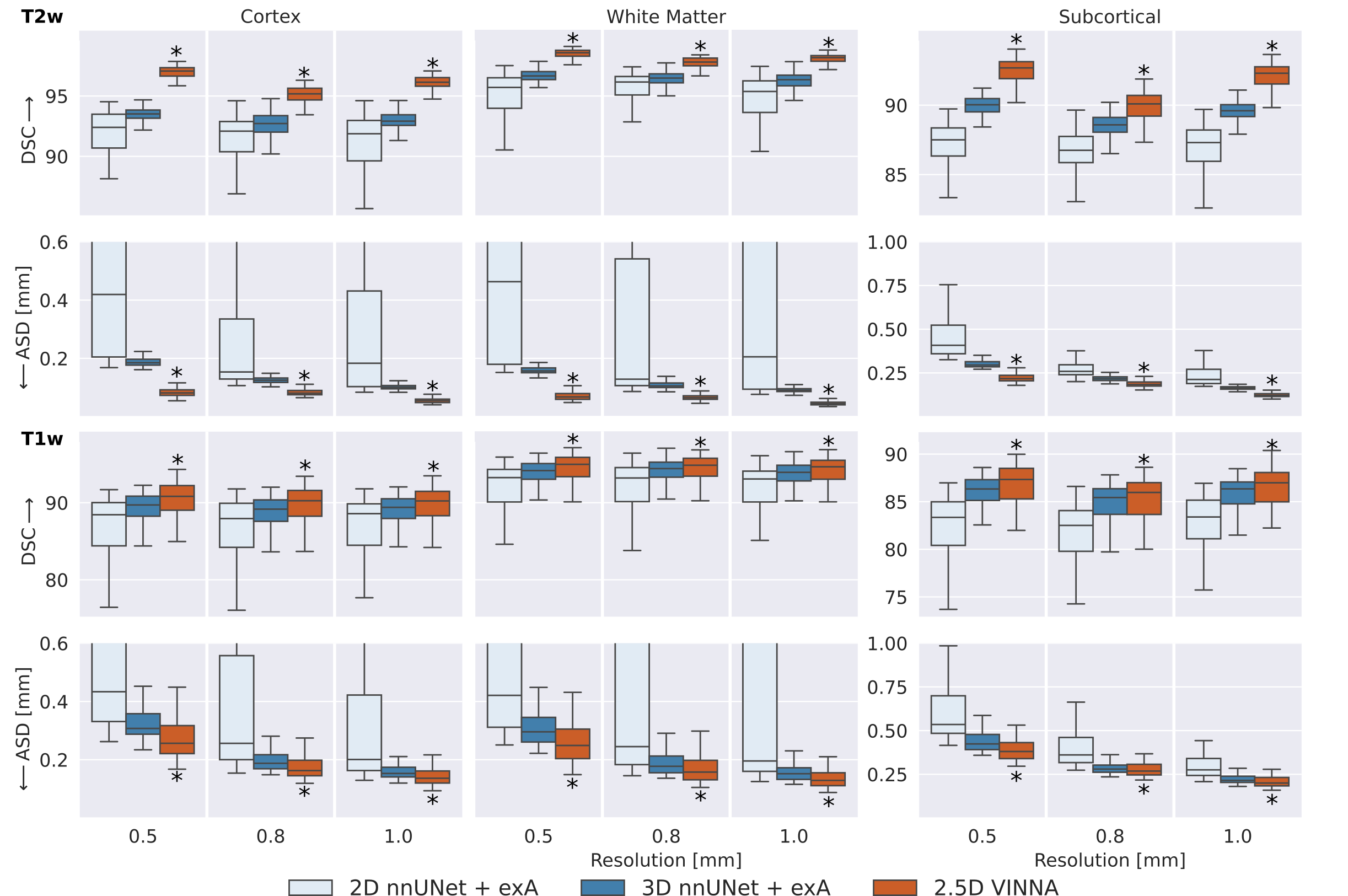}
    \caption{SOTA Segmentation performance: Our \ac{VINNA} with the 4-DOF transform module (last box in group) outperforms the three state-of-the-art deep-learning approaches, 2D nnUNet + exA and 3D (first and second box) in Dice Similarity Coefficient (DSC, upper row) and average surface distance (ASD, lower row). This performance advantage is significant (corrected $p < 10^{-10}$, indicated with *) and consistent across three resolutions (\SI{0.5}{\milli\meter}, \SI{0.8}{\milli\meter}, and \SI{1.0}{\milli\meter}), two modalities (T2w, top and T1w, bottom), and three structure groups (cortex, \ac{WM} and subcortical structures).} 
    \label{fig:sota_res}
\end{figure*}

\Cref{fig:sota_res} shows a detailed comparison of three different deep-learning based methods for neonate segmentation across modalities (T2w top, T1w bottom) and resolutions. \new{All models are trained under equal parameter and dataset settings.}
Comparing performance between the two modalities shows, that all models perform better on the T2w (top) than the T1w \acp{MRI} (bottom) across all structures. With \ac{VINNA} (right box in each plot), the reduction is similar across all resolutions, with an average difference in \ac{DSC} of 5.85, 3.47, and 5.16 on the cortex, \ac{WM}, and subcortical structures. The \ac{ASD} is, on average, improved by \SI{0.09}{\milli\meter} when predicting on the T2w instead of T1w inputs. The nnUNet framework in 2D (left box) and 3D (second from left) has less improvement on the T2w images with an average difference between T1w and T2w of 3.75, 2.44 and 3.70 in \ac{DSC} and \SI{0.04}{\milli\meter} \ac{ASD} on the aforementioned structures.

When comparing the four models, the 2D nnUNet + exA (left box) version performs worse than the 3D (second from left box), and 2.5D \ac{VINNA} (right box) across all resolutions, structures, and modalities. Particularly notable are the large variations of 2D nnUNet + exA in prediction performance (large standard deviation) and large \ac{ASD} (see \Cref{fig:sota_res}), especially at the highest resolution (\SI{0.42}{\milli\meter}, for the cortex, \SI{0.46}{\milli\meter} for \ac{WM} and \SI{0.41}{\milli\meter} for subcortical structures). This difference is less prominent in the \ac{DSC} scores, but 2D nnUNet + exA also performs worst across all resolutions with respect to this metric (i.e.\ 92.39, 95.70, and 87.50 for a resolution of \SI{0.5}{\milli\meter}). The 3D nnUNet + exA (second from left) improves accuracy by 24.5~\%, 12.3~\%, and 22.9~\% for \ac{ASD} and 1.75~\%, 1.06~\%, and 1.64~\% for \ac{DSC} across the three different resolutions.  \ac{VINNA} with its 4-DOF transform module (\ac{VINNA}, right box) is again the best performing model, significantly outperforming all other networks. Compared to the 3D nnUNet + exA, \ac{ASD} and \ac{DSC} scores are significantly improved with the highest gain on the cortical structures (56~\%, 35~\%, and 45~\% \ac{ASD} and 3.8~\%, 2.7~\%, and 3.5~\% \ac{DSC} for \SI{0.5}{\milli\meter}, \SI{0.8}{\milli\meter} and \SI{1.0}{\milli\meter}, respectively).

\begin{figure*}[!tb]
    \centering
    \includegraphics[width=\textwidth,keepaspectratio]{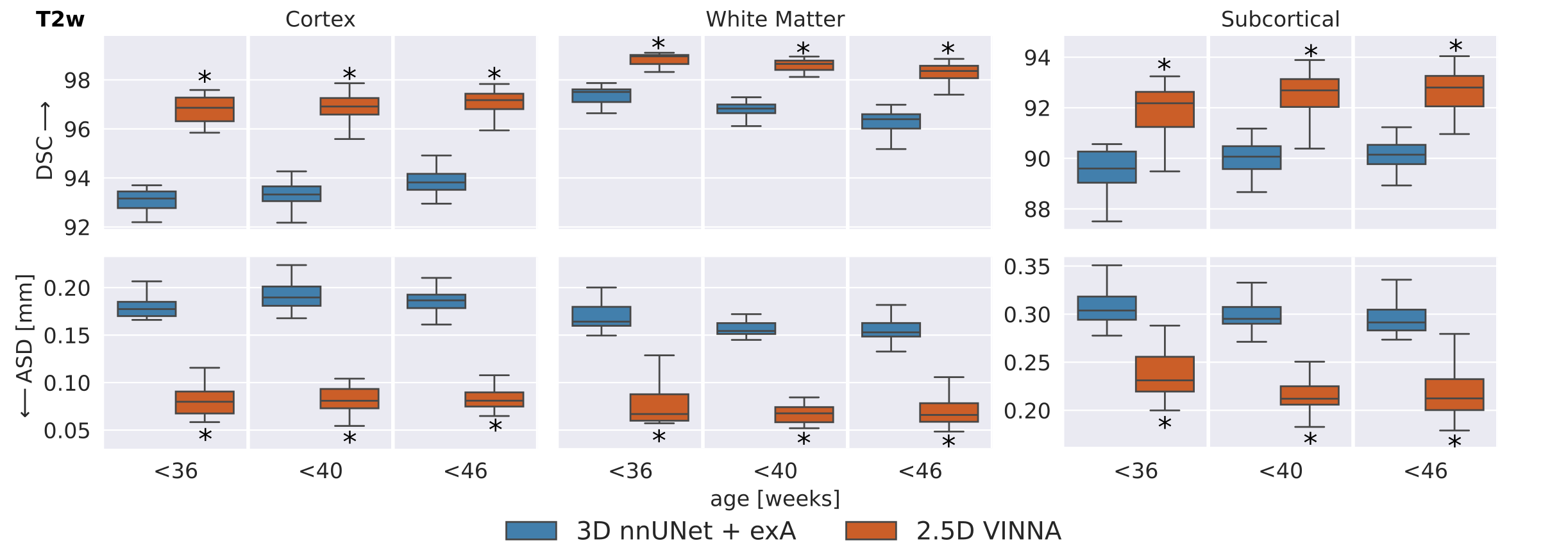}
    \caption{SOTA Segmentation performance across age groups: \ac{VINNA} equipped with the 4-DOF transform module (right box) consistently outperform state-of-the-art deep-learning approaches 3D nnUNet with external augmentation (+exA, left box) across four age groups. Improvement in dice similarity coefficient (DSC, top) and average surface distance (ASD, bottom) on T2w at \SI{0.5}{\milli\meter} is significant (corrected $p < 10^{-4}$, indicated with *) for $<36$, $<40$ and $<46$ week old newborns. } 
    \label{fig:sota_age}
\end{figure*}

To evaluate if this trend is consistent across age groups, the \SI{0.5}{\milli\meter} \ac{MRI} images are split into three approximately equal sized groups based on the participants' age-at-scan information (32-36, 36-40 and 40-46 weeks). \Cref{fig:sota_res} shows \ac{DSC} (top) and \ac{ASD} (bottom) calculated for T2w (\Cref{fig:sota_age}) for the 3D nnUNet + exA (left box) and \ac{VINNA} with its 4-DOF transform module (right box) in each of the categories.

Consistent with the previous section, 3D nnUNet + exA reaches the weakest \ac{ASD} and \ac{DSC} across all age groups. On average, the \ac{VINNA} with its 4-DOF transform module (right box) improves performance compared to nnUNet by 2.8, 2.9 and 2.8~\% \ac{DSC} and 41.5, 43.6 and 43.1~\% \ac{ASD} from the youngest (32-36) to the oldest ($< 46$) age group and reaches a \ac{DSC} of 96.86, 96.913, 97.17 for the cortical structures, 98.96, 98.652, 98.362 for the \ac{WM} structures and 92.18, 92.69, 92.80 for the subcortical structures across all age groups (youngest to oldest) on the T2w \acp{MRI} (\Cref{fig:sota_age}, top). Here, \ac{VINNA} also reaches the lowest \ac{ASD} (\SI{0.080}{\milli\meter}, \SI{0.081}{\milli\meter}, \SI{0.081}{\milli\meter} for the cortical structures, \SI{0.067}{\milli\meter}, \SI{0.068}{\milli\meter}, \SI{0.066}{\milli\meter}, for the \ac{WM} structures and \SI{0.231}{\milli\meter}, \SI{0.212}{\milli\meter}, \SI{0.212}{\milli\meter}, for the subcortical structures). The results are significantly better compared to 3D nnUNet + exA (corrected $p < 10^{-9}$). 
As seen by the increasing \ac{DSC} and decreasing \ac{ASD}, the younger age groups ($<32-36$) have proved to be more challenging to segment. For \ac{VINNA}, the performance decreases most significantly for the subcortical structures (\ac{DSC} by 1.82~\% and \ac{ASD} by 20.09~\%) and least on the cortex (0.35~\% \ac{DSC} and 4.65~\% \ac{ASD}). This trend is consistent for the other two models. 
Assessment of qualitative differences between the 3D nnUNet + exA and \ac{VINNA} on a representative participant at 40 weeks of age (\Cref{fig:qual}) shows slight over-segmentation of the cortex and loss of small \ac{WM} strands with the 3D nnUNet + exA (third row, second column, arrows). Overall, the segmentation with \ac{VINNA} (fourth row) appears less smoothed and closer to the ground truth (second row). 

\subsubsection{Comparison to \ac{iBEAT}}
\begin{figure*}[!tb]
    \centering
    \includegraphics[width=\textwidth,keepaspectratio]{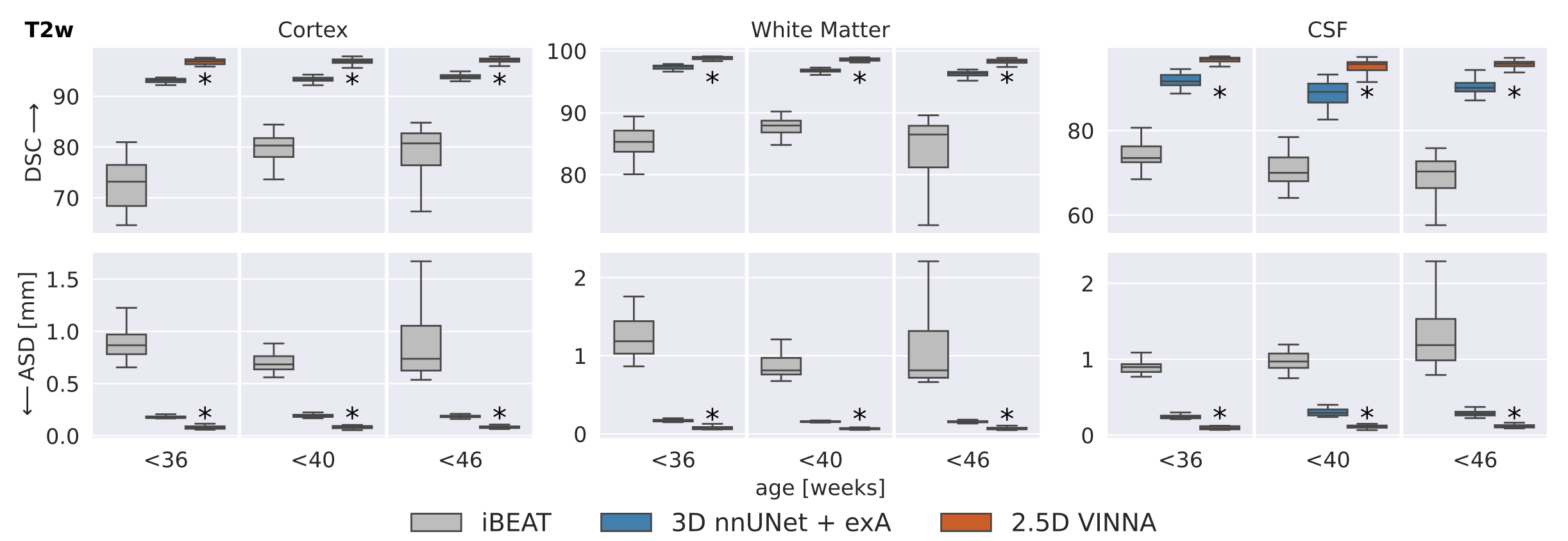}
    \caption{Deep-Learning networks versus \ac{iBEAT}: Similarity to the \ac{dHCP} reference is higher with \ac{VINNA} (third box) and 3D nnUNet + exA (second box) than \ac{iBEAT} (first box) with respect to dice similarity coefficient (DSC, top) and average surface distance (ASD, bottom) on T2w \acp{MRI} at 0.5 mm across three age groups. Segmentation results with \ac{VINNA} are significantly closer to \ac{dHCP} (corrected $p < 10^{-6}$, indicated with *) for \ac{CSF}, \ac{GM}, and \ac{WM}. Note, that \ac{iBEAT}´s structure definition is not identical to the \ac{dHCP}-ALBERTs atlas and analysis is based on harmonized, merged labels. } 
    \label{fig:ibeat}
\end{figure*}

In \Cref{fig:ibeat}, the deep-learning models are compared to the docker version of \ac{iBEAT} on the T2w images at \SI{0.5}{\milli\meter}. \ac{iBEAT} is officially designed for ages 0-6 years and the docker version we used for processing returns three labels (\ac{WM}, \ac{GM}, \ac{CSF}). The definition of these labels is different from that of the \ac{dHCP}-atlas (see \Cref{sec:label-harm}), so we map the ground truth as well as the predictions from the deep-learning networks (nnUNet3D and \ac{VINNA}) to be able to compare segmentation similarity. As described in \Cref{sec:traditional-tools}, retraining the \ac{CNN} part of \ac{iBEAT} under same data and label definitions is not possible, as neither the source code nor the  original training data is available online. Note, even though we do not need to interpolate, cross-protocol comparisons include atlas differences and may introduce additional errors due to the mapping. While results should be interpreted with the caveat that \ac{iBEAT} uses a different atlas and training dataset than 3D nnUNet + exA and \ac{VINNA}, the label harmonization allows an as-fair-as-possible comparison with this state-of-the-art method. 

With respect to the mapped \ac{dHCP}-reference segmentation, \ac{DSC} (top) and \ac{ASD} (bottom) are lower for \ac{iBEAT} (left box in each plot) compared to the the deep-learning methods (3D nnUNet + exA, middle box, and \ac{VINNA}, right box in each plot). Performance of \ac{iBEAT} on the \ac{GM}, \ac{WM}, and \ac{CSF} improves with age. \ac{GM} and \ac{WM} are closest to the reference at 36-40 weeks (\ac{DSC} 80.29/87.94 and \ac{ASD} \SI{0.684}/\SI{0.813}{\milli\meter}) while \ac{CSF} peaks at 32-26 weeks (73.49 and \SI{0.896}{\milli\meter}). The 3D nnUNet + exA and \ac{VINNA} show a similar trend, but performance is more consistent. 
As mentioned before, the differences are not necessarily due to wrong predictions made by \ac{iBEAT}. Looking at the qualitative comparison in \Cref{fig:qual}, differences appear small, with \ac{iBEAT} (third row) missing a few \ac{WM} strands (arrow) and slightly over-segmenting the cortex compared to the mapped ground truth (second row).

\subsubsection{Comparison to \ac{infantFS}}
\begin{figure*}[!tb]
    \centering
    \includegraphics[width=\textwidth,keepaspectratio]{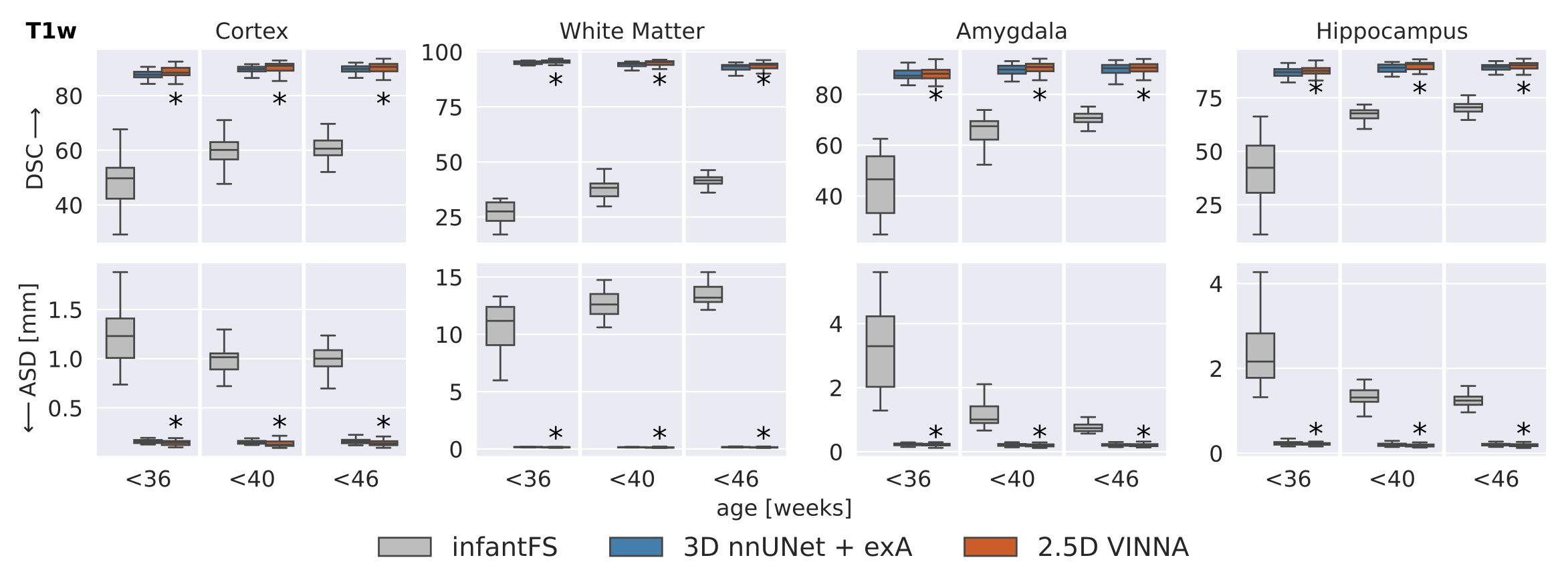}
    \caption{Deep-learning networks versus \ac{infantFS}: \ac{VINNA} with the 4-DOF transform module (third box) and 3D nnUNet + exA (second box) are closer to the \ac{dHCP} reference than \ac{infantFS} (first box) on T1w at \SI{1.0}{\milli\meter}, the supported modality and resolution for \ac{infantFS}. Dice Similarity Coefficient (DSC, top) and average surface distance (ASD, bottom) significantly improve with \ac{VINNA} (corrected $p < 10^{-6}$, indicated with *) on the cortex, \ac{WM}, hippocampus and amygdala across all age groups. Note, that definition of subcortical structures differs in \ac{infantFS} and predictions are harmonized to allow comparison to the deep-learning models.} 
    \label{fig:infantfs}
\end{figure*}

\Cref{fig:infantfs} shows performance comparison between \ac{infantFS} (left box) and the deep-learning methods, 3D nnUNet + exA (middle box) and \ac{VINNA} (left box), but in contrast to previous evaluations on T1w images at 1.0 mm, the operating resolution and modality of \ac{infantFS}. Note, that the \ac{infantFS} labels are also different from those of \ac{dHCP} and the ground truth labels must be mapped (see \Cref{sec:label-harm} for details). As for \ac{iBEAT}, the cross-protocol comparison can put the traditional method at an unfair disadvantage and results should be considered with this caveat. 

\begin{figure}[!tb]
    \centering
    \includegraphics[width=\columnwidth,keepaspectratio]{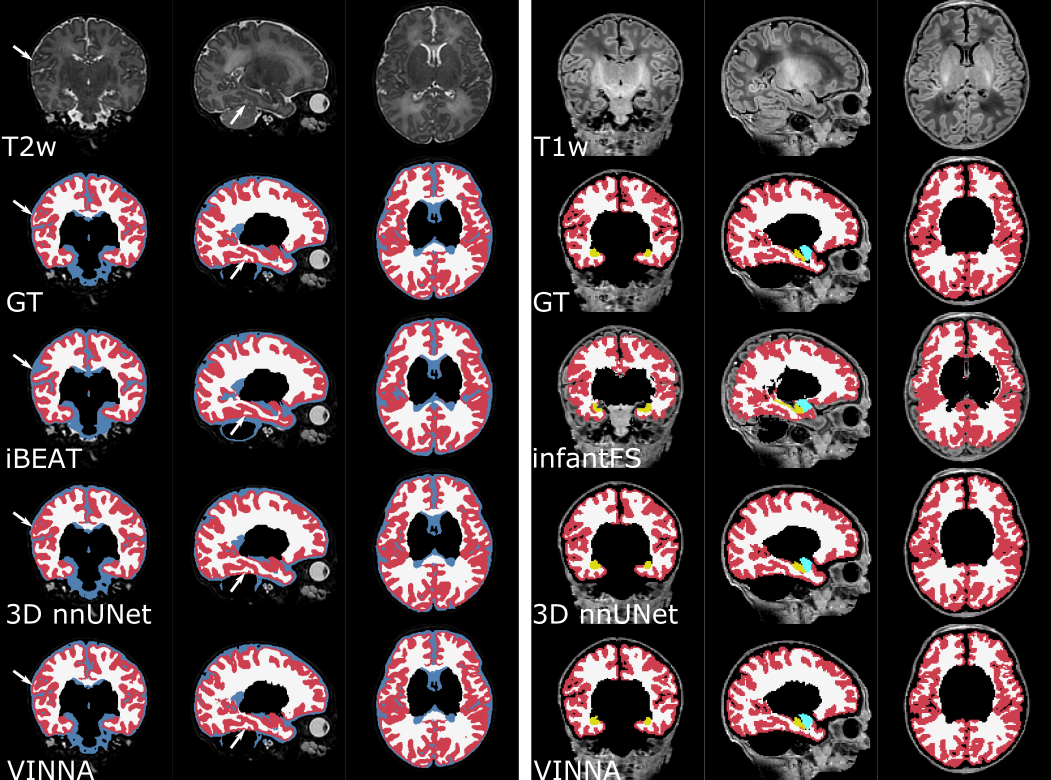}
    \caption{Qualitative T1w and T2w \ac{MRI} segmentations on a representative scan at 41 weeks. \ac{VINNA} with the 4-DOF transform module (last row) captures structural details lost in other methods. Comparison of the mapped ground truth (top) and segmentations from \ac{iBEAT} (third row), 3D nnUNet + exA (fourth row) on a representative participants T2w \ac{MRI} (left). The right side shows the T1w-scan from the same participant at 1.0 mm with ground truth (top), \ac{infantFS} (third row) and the deep-learning methods. } 
    \label{fig:qual}
\end{figure}

Overall, the \ac{infantFS} predictions differ strongly from the mapped \ac{dHCP} ground truth, specifically for the younger age ranges. The highest similarity is reached for the subcortical structures (amygdala and hippocampus) at 40-46 weeks (\ac{DSC} of 70.72/70.56 and \ac{ASD} of \SI{0.736}/\SI{1.23}{\milli\meter}, respectively). The cortex and \ac{WM} reach a maximum \ac{DSC} of 60.60/41.62 and \ac{ASD} of \SI{1.0}/\SI{13.20}{\milli\meter}. Qualitative comparison (see \Cref{fig:qual}, third row, left panel) shows difficulties with the correct location of the \ac{GM} and \ac{WM} border on the \ac{dHCP} T1w MRI. Larger portions of the cortex are under-segmented and strands of \ac{WM} are lost compared to the ground truth. The deep-learning methods reach a \ac{DSC} above 80 and an \ac{ASD} below \SI{0.5}{\milli\meter} for all structures and age groups. The method closest to the \ac{dHCP} reference is again \ac{VINNA} with the 4-DOF transform module (left box), followed by 3D nnUNet + exA (middle box). For a detailed comparison between nnUNet and \ac{VINNA} on T1w, see \Cref{fig:sota_age}.

\subsection{External validation on manual labels (M-CRIB)}\label{sec:results-generalization}
To assess generalizability to a different dataset in our target age range (20-40 weeks) and to provide results with respect to a manual reference, we compare the segmentations produced by \ac{VINNA}, 3D nnUNet + exA, and the dhcp-minimal-processing-pipeline to the \SI{0.62}{\milli\meter} high-resolution T2w scans forming the M-CRIB atlas \cite{mcrib_2019}. This dataset contains T2w \acp{MRI} from 10 participants and accompanying label maps based on the Desikian-Killiany-Tourville (DKT) atlas \cite{Klein2012}. Note, that the labels are not identical to the \ac{dHCP}-ALBERTs atlas. Hence, we combine the cortical parcels to one label (cortex) for the segmentation comparison and mask all but three structures (\ac{WM}, hippocampus, and lateral ventricles). As \ac{iBEAT} does not differentiate between subcortical structures and \ac{GM} or \ac{WM}, nor \ac{CSF} and ventricles, mapping of both, the ground truth and prediction, would be different compared to nnUNet, \ac{VINNA}, and the dhcp-minimal-processing-pipeline. A fair comparison is therefore only possible between the latter methods, and \ac{iBEAT} is thus not included in the following section.

\begin{figure}[!tb]
    \centering
    \includegraphics[width=\columnwidth,keepaspectratio]{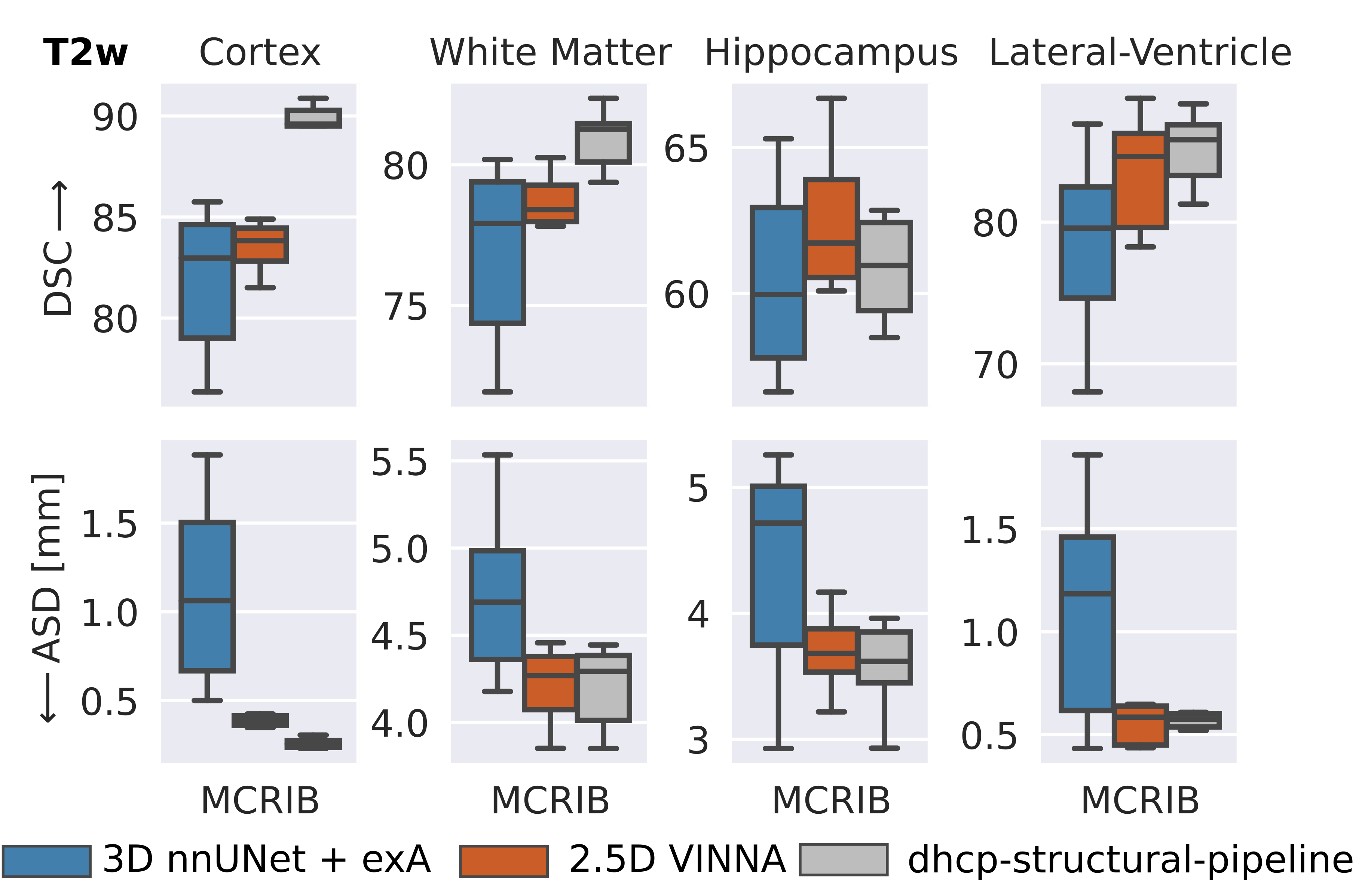}
    \caption{External validation of segmentation performance on the M-CRIB dataset. \ac{VINNA} (second box) outperforms nnUNet3D (first box) on all four structures and the dhcp-strucutral-pipeline (third box) on the hippocampus. Dice Similarity Coefficient (DSC) and average surface distance (ASD) for cortex, \ac{WM}, hippocampus and lateral-ventricles are calculated with respect to harmonized manual labels on ten subjects from M-CRIB. Note, that atlas definition differs between ground truth and predictions.} 
    \label{fig:mcrib}
\end{figure}

In \Cref{fig:mcrib}, the \ac{DSC} (top) and \ac{ASD} (bottom) are compared over four structures (from left to right: cortex, \ac{WM}, hippocampus and lateral-ventricles) for the deep-learning methods (3D nnUNet + exA and \ac{VINNA}) as well as the dhcp-minimal-processing-pipeline. Predictions for these methods are all based on the same label definition (\ac{dHCP}-ALBERTs atlas \cite{gousias-2012,makropoulos-2014}). As for the \ac{dHCP} test set, \ac{VINNA} outperforms the 3D versions of nnUNet across all four structures with a \ac{DSC} of 83.83 and an \ac{ASD} of \SI{0.387}{\milli\meter} on the cortex, 78.41 and \SI{4.268}{\milli\meter} on the \ac{WM}, 61.73 and \SI{3.682}{\milli\meter} on the hippocampus, and 84.62 and \SI{0.585}{\milli\meter} on the lateral ventricles. Compared to nnUNet, the performance improves on average by 3.36~\% for the \ac{DSC} and 19.40~\% for the \ac{ASD}. Furthermore, 3D nnUNet + exA incorrectly flips left-right labels for five participants (P02-P04, P08-P09). We restore the lateralization in the presented results as \ac{DSC} and \ac{ASD} would have otherwise been close to zero for half of the participants. 
On the hippocampus, ventricles, and \ac{WM} performance of \ac{VINNA} is similar to the dhcp-minimal-processing-pipeline, which is the best performing method. Specifically, predictions of the cortex are closer to the M-CRIB manual labels with the dhcp-pipeline (\ac{DSC} of 89.60 and \ac{ASD} of \SI{0.257}{\milli\meter}), which is likely due to the reliance on surface models, while the trained deep-learning models seem to over-segment the cortex (indicated by the similarity of the \ac{ASD}, but larger differences in the \ac{DSC}). Note that \ac{VINNA} performs slightly better on the hippocampus with an increase in \ac{DSC} by 1.25~\% and \ac{ASD} by 0.58~\%. Due to the low number of participants, significance tests are not applicable for these experiments.



\section{Discussion}

In this paper, we present \ac{VINNA} -- a resolution-independent network for native-resolution neonatal brain \ac{MRI} segmentation. With \ac{VINNA} and our novel network-integrated 4-DOF transform module, we address two main difficulties associated with neonate segmentation: resolution non-uniformity across data cohorts and the extended range of head positions in infants. 

In contrast to adults, newborn head positioning in the scanner varies significantly due to imaging during sleep, smaller head sizes, and relevant necessary modifications to scanner equipment, such as the padding of head coils. Additionally, while scans are commonly recorded at high resolutions, no uniform standard exists across cohorts. The availability of newborn datasets is also scarcer than that of adult subjects and the existing collections to date are unlikely to represent that wide diversity in resolutions and head positions.  


The current state-of-the-art to address spatial variability such as head positions is data augmentation, which applies randomly sampled scale, rotation, and translation transformations in the native imaging space (i.e., externally to both intensity and label map). In \ac{VINNA}, we introduce the 4-DOF transform module that can apply such transformations internally as part of the network. As the parameters to the transformation are inputs to the network, they can be randomized during training, similarly to external data augmentation methods. Moving the augmentation operation into the network, so it acts upon feature maps instead of inputs, marks a paradigm shift for data augmentation strategies. While we only implemented an augmentation of 4 DOFs here, the concept may be generalized to 9 DOFs for 3D or even to warp fields as well as to other tasks such as classification, regression, etc.

We demonstrate that the new network-integrated 4-DOF transform with internal augmentation outperforms state-of-the-art external augmentation approaches in \ac{CNN}*, \ac{VINN}, and nnUNet \cite{isensee_2021} on the \ac{dHCP} cohort. Across three different resolutions and two modalities, our \ac{VINNA} achieves the highest \ac{DSC} (95.33 on average), as well as the lowest \ac{ASD} (\SI{0.102}{\milli\meter} on average). Metric evaluation combined with qualitative inspection indicate that the internal 4-DOF transform module in \ac{VINNA} better retains high-level details across all resolutions and age groups. 

To better explain the factors and mechanisms driving the performance improvements of \ac{VINNA}, we review the observation from FastSurferVINN \cite{Henschel_2022} that motivated the extension presented here: in one-to-one comparisons, external augmentation reduces the segmentation performance on sub-millimeter \acp{MRI}. While -- at first sight -- the addition of data augmentation reducing performance seems contradictory, the one-to-one comparison of VINNA and VINNA + exA (see \Cref{fig:ablation}, the only difference is added external augmentation) robustly confirms the observation and extends it from just scaling to rigid transforms.

The positive effect of data augmentation is usually associated with an expansion of the input dataset through equivariant operations. Implementing operations for image and label pairs that are truly equivariant can be difficult. We believe that the loss of information due to image interpolation (lossy interpolation of the label map and image) is larger than previously believed. Internal augmentation, for the first time, offers an alternative approach with interpolation of continuous values in multiple feature maps, reducing the information loss. 

Furthermore, the 4-DOF transform module together with the internal augmentation regularizes the latent space of the network, because it imposes an additional constraint: spatial consistency of the feature maps. Compared to equivalent CNN architectures, VINNA (and VINN) also benefit from a reduced capacity requirement to capture a large range of resolutions.

\new{Our comparison to nnUNet highlights, that 2D approaches lack contextual information, and fail to provide reliable predictions for whole brain segmentation. The compromise between mid-range and long-range context in the 2.5D \ac{VINNA} recovers structural information better and achieves higher segmentation performance across all age groups and structures - even compared to 3D methods. As full-view 3D networks are currently not applicable for high-resolution \ac{MRI} segmentation due to memory requirements, nnUNet and other 3D networks rely on patch-based processing. In this case, the increased 3D context comes at the cost of limited long-range information and features a smaller field of view, potentially explaining the observed reduction in accuracy compared to 2.5D networks. This finding is in line with previous investigations which found limited performance differences between 2.5D and 3D approaches, even after extensive optimization of the 3D network architectures \cite{roy2D3Dcomp2022}.}

On the \ac{dHCP} cohort, \ac{VINNA} and its 4-DOF transform module also emulates the segmented structures better than traditional state-of-the-art infant pipelines, namely \ac{infantFS}\cite{infantfs} and \ac{iBEAT}\cite{wang_ibeat_2023}. Notably, \ac{infantFS} relies on traditional atlas-based segmentation while \ac{iBEAT} uses a combination of deep-learning and traditional tools with a number of \acp{CNN} trained on defined target age-ranges. While the re-trained networks (nnUNet + exA and VINNA) reach better results with respect to \ac{DSC} and \ac{ASD}, it should  be noted that both, \ac{infantFS} and \ac{iBEAT}, differ significantly with respect to the returned number and definition of segmented regions. The necessary mapping between the segmentations is bound to introduce a bias, which can not be easily assessed. Additionally, both pipelines cater to a slightly different, larger age range (0-2 years for \ac{infantFS} and 0-6 years for \ac{iBEAT}). Consequently, predictions from both methods of the cortex and \ac{WM} improve for participants closer to the officially supported age-range ($>$ 40 weeks). Qualitative assessment also shows good performance for the older newborns in \ac{iBEAT}. \ac{infantFS} unfortunately fails to correctly capture the cortex and \ac{WM} on the majority of participants.  
The original \ac{iBEAT} v2.0 paper\cite{wang_ibeat_2023} also evaluates performance on the \ac{dHCP} data and reports a \ac{DSC} of ~0.9 for the \ac{WM} and 0.85 for the \ac{GM}. Our results are in concordance with this assessment for the $>$ 40 weeks old participants (\ac{DSC} of 0.83 on the \ac{GM} and 0.88 on the \ac{WM}). \new{The authors do not provide information on their label harmonization, therefore we can not infer their reference standard. In contrast to the docker v2.0 version, the cloud version of \ac{iBEAT} (not available for local data processing) does provide cortical parcellations. Extracting the cortex from the \ac{GM} label (a combination of both cortical and subcortical \ac{GM} in the docker version) allows direct comparison to the \ac{dHCP} solution after merging its cortical structures, possibly explaining performance differences.}. In summary, \ac{iBEAT} seems to work well on the supported age range while \ac{infantFS} is less precise on the \ac{dHCP} population. Other features included in the pipelines, such as surface generations are an advantage compared to the proposed \ac{VINNA} and can help to refine the segmentation of convoluted structures such as the cortex \cite{fischl2012freesurfer}. For our target domain in this paper, however, the \ac{VINNA} architecture appears to emulate the investigated tissue classes more precisely.

Due to the limited extrapolation capabilities of neural networks, generalizability beyond the training set distribution is, however, uncertain. While the 4-DOF transform module in \ac{VINNA} serves as a diversification of the training distribution with respect to spatial orientations and image resolution, the base cohort is still only a representation of the \ac{dHCP} population, i.e.\ all scans encountered during training represent newborns between 20-40 weeks post-gestational age from a control cohort acquired on the same 3~T Phillips scanner. Therefore, dedicated experimental validation is required to confirm the models' effectiveness under differing conditions. As for all automated methods, manual quality checks of the predictions are recommended. 
While \ac{VINNA} does perform well on M-CRIB, which covers the same age range as the \ac{dHCP}, generalization to other cohorts is not necessarily guaranteed. Specifically, the T1w image intensities in \ac{dHCP} appear significantly different from other cohorts which might also explain why \ac{infantFS} performs poorly on the testing set.  The T2w \acp{MRI} in \ac{dHCP} are, on average, of better quality and the dhcp-minimal-processing-pipeline builds the ground truth segmentations based on it \cite{dhcp_pipeline2018}. 
Additionally, in the early weeks of life, tissue contrast is higher in T2w recordings as the brain is not fully matured and myelination is still ongoing \cite{dubois_2021,miller2012}. Structural details and specifically tissue boundaries might be missing, blurred, or ambiguous in T1w \ac{MRI}. Hence, the imaging data may lack sufficient information to allow correct delineation of the (sub-)cortical structures. This may also explain why the deep-learning networks (i.e.\ nnUNet, \ac{CNN}*, \ac{VINN}), and \ac{VINNA} are not able to emulate the ground truth on the T1w \ac{MRI} as closely as on the T2w images.  

Better accessibility of newborn datasets would allow diversification of the training sets and subsequently a better representation of the newborn \ac{MRI} distribution with respect to both, T1w and T2w modalities. It has been shown that an increase in the training corpus alone is extremely effective to boost performance \cite{Sun2017, Henschel_2022}. Age specific models, as our \ac{VINNA} or \ac{iBEAT}´s \acp{CNN}, are another way to reduce variations and therefore, segregate the problem (i.e.\ less variations within one age group). However, the limited data availability and non-uniform segmentation labels still pose significant barriers. Models for more specific segmentations than just \ac{CSF}, \ac{WM} and \ac{GM} are currently not trainable in a supervised fashion due to missing ground truth. In addition, definition of these three structures alone already varies across different atlases and tools, which makes fair method comparisons challenging. Neither manual labels nor automated segmentation tools exist for a unified segmentation definition across different resolutions, modalities and age-ranges. The M-CRIB atlas \cite{mcrib_2019}, an infant-specific version of the Desikan-Killiany-Tourville-Atlas \cite{Klein2012} that is commonly used in adults, provides a first step towards this goal. A consistent structure definition across different stages of life is especially important in the context of longitudinal studies, as segmentation with age-dependent models can induce biases and reduce anatomical consistency \cite{reuter2012}. How to solve this conundrum is an open question for the future. Several NIH- and internationally-funded initiatives have recently been dedicated to acquire data from newborns \cite{dhcp_pipeline2018}, infants and pediatric age ranges \cite{bcp,hbcd,Ping2016} as well as adolescence \cite{karcher_abcd_2021}. Due to the easy integration of varying data resolutions and accommodation for head position variations between infants, toddlers, and adults, our \ac{VINNA} architecture might prove to be useful in this area once the data and label availability problem is resolved.

Overall, with \ac{VINNA}, we provide a fast and accurate method for high-resolution subcortical structure segmentation, cortical and \ac{WM} parcellation of neonatal T1w and T2w \ac{MRI} which generalizes well across the \ac{dHCP} cohort. The presented 4-DOF transform module is also easy to integrate into other network architectures and might prove useful in different areas dealing with strong orientation variations. The application to neonates will be made available under VINNA4neonates as an open source package\footnote{on our github pages}. Adaptation of the  \ac{infantFS} surface pipeline to the VINNA4neonates predictions, similar to the approach taken in FastSurfer\cite{Henschel_2020} for adults, is an exciting direction for future work.

\section{Data and Code Availability}
All \ac{MRI} datasets used within this article are publicly available and the open source repositories are cited within the article (\Cref{sec:datasets}). 

The source code of VINNA4neonates will be made publicly available on Github (\url{https://github.com/deep-mi/NeonateVINNA}) upon acceptance. 

\section{Author Contributions}
Leonie Henschel: Conceptualization, Methodology, Software, Validation, Formal analysis, Investigation, Writing - original draft, Writing - review \& editing, Visualization.

David Kügler: Conceptualization, Methodology, Software, Writing - review \& editing, Supervision. 

Martin Reuter: Conceptualization, Methodology, Resources, Writing - review \& editing, Supervision, Project administration, Funding acquisition.

Lilla Zöllei: Conceptualization, Methodology, Resources, Writing - review \& editing, Supervision, Project administration, Funding acquisition.

\section{Competing Interests}
The authors declare that they have no known competing financial interests or personal relationships that could have appeared to influence the work reported in this paper.

\section{Acknowledgements and Funding}

This work was supported by DZNE institutional funds, by the Federal Ministry of Education and Research of Germany (031L0206, 01GQ1801), by NIH (R01 LM012719, R01 AG064027, R56 MH121426, P41 EB030006, and 5R01 HD109436-01), NICHD, and a NIH Shared Instrument Grant (S10RR023043). Funding was further received from the Chan-Zuckerberg Initiatives Essential Open Source Software for Science RFA (EOSS5 2022-252594).

Data were provided by the Melbourne Children´s Regional Infant Brain Atlas and the Developing Human Connectome Project, KCL-Imperial-Oxford Consortium funded by the European Research Council under the European Union Seventh Framework Programme (FP/2007-2013) / ERC Grant Agreement no. [319456]. We are grateful to the families who generously supported this trial.

\bibliographystyle{elsarticle-num}
\bibliography{mybibfile}

\section*{Appendix}
\setcounter{table}{0}
\renewcommand\thetable{\Alph{table}}
\setcounter{equation}{0}
\renewcommand\theequation{\Alph{equation}}
\setcounter{section}{1}
\renewcommand\thesection{\Alph{section}}

\subsection{Ablation on T1w}\label{sec:appendix_ablation_t1}

\begin{figure*}[h!]
    \centering
    \includegraphics[width=\textwidth,keepaspectratio]{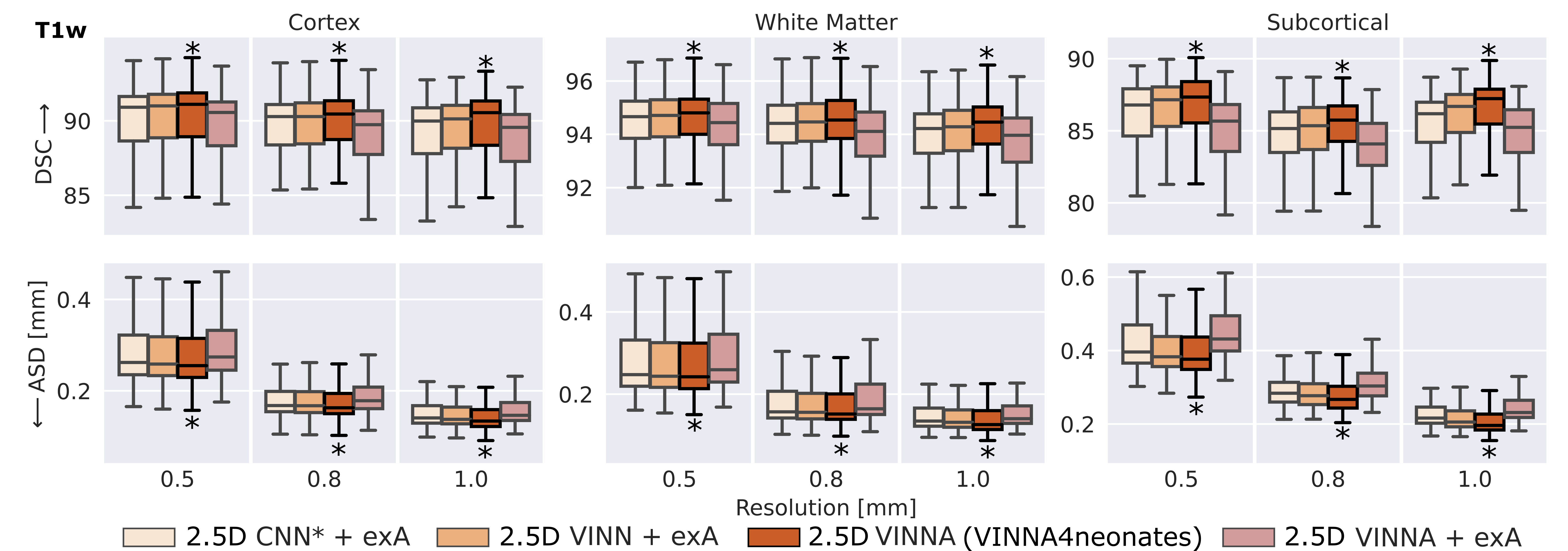}
    \caption{Comparison of approaches for external and internal spatial augmentation on T1w images: Our VINNA method -- with the 4-DOF transform module -- (third box in group) outperforms both state-of-the-art approaches with external Augmentation (CNN* + exA and VINN + exA, first and second box) in Dice Similarity Coefficient (DSC, upper row) and average surface distance (ASD, lower row). This performance advantage is significant (corrected $p < 0.05$) and consistent across resolutions and structures. Combining VINNA with external augmentation (VINNA + exA, last box) reduces performance. 
    exA: external Augmentation.} 
    \label{fig:ablation_t1}
\end{figure*}

\Cref{fig:ablation_t1} shows \ac{DSC} (top) and \ac{ASD} (bottom) for the ablative evaluation external augmentation (exA) versus the internal 4-DOF transform modul (\ac{VINNA}) on the T1w. CNN* + exA (first box) reaches the weakest results, followed by \ac{VINN} + exA (second box) and \ac{VINNA} (third box). The performance improves the most on the subcortical structures (2.44 \% \ac{ASD} and 0.4 \% \ac{DSC} when switching from CNN* to \ac{VINN}, 5.91 \% \ac{ASD} and 0.85 \% \ac{ASD} when switching to \ac{VINNA}). Performance improvements with the 4-DOF transform module in \ac{VINNA}, is significant (corrected $p < 0.05$) and results in an average final DSC of 90.95, 90.03 and 90.36 and ASD of \SI{0.295}{\milli\meter}, \SI{0.2}{\milli\meter}, \SI{0.2}{\milli\meter} for the \SI{0.5}{\milli\meter}. \SI{0.8}{\milli\meter}, and \SI{1.0}{\milli\meter} resolution, respectively. Re-addition of exA to \ac{VINNA} (fourth box) reduces performance to a similar level as CNN* + exA.

Overall, all models perform better on T2w (\Cref{fig:ablation}) rather than T1w \acp{MRI} (\Cref{fig:ablation_t1}) across all structures. When segmenting a T2w \ac{MRI} of the same individual, peformance improves for all networks, on average, by 5.7, 3.63 and 5.09 \ac{DSC} points for the cortex, \ac{WM}, and subcortical structures, respectively. The \ac{ASD} difference between T1w and T2w applications nearly doubles for the highest resolution (\SI{0.5}{\milli\meter}) with an average of \SI{0.18}{\milli\meter} compared to both lower resolutions (\ac{ASD} of \SI{0.9}{\milli\meter} for \SI{1.0} and \SI{0.8}{\milli\meter}) for the three structure averages. The \ac{DSC} point improvement for the T2w images is similar across all resolutions. The trend is consistent for both networks (CNN* and \ac{VINN}) and augmentation schemes (external versus internal).

\vspace{-2ex}
\subsection{Dataset Summary}
\vspace{-4ex}
\begin{table}[h!]
\centering
\caption{Summary of datasets used for training, validation, and testing. Table lists the dataset, number of participants (\#P), range of post-gestational age at scan in weeks and isotropic resolution in millimeter (Res [mm]) used in the paper.  }
\begin{tabular}{c|c|c|c|c}
\textbf{Usage}            & \textbf{Dataset}               & \textbf{\#P}               & \textbf{\begin{tabular}[c]{@{}c@{}}Age \\ {[}weeks{]}\end{tabular}} & \textbf{Res {[}mm{]}}                 \\ \hline
Training                  & dHCP                           & 320                        & 26-44                                                               & 0.5, 0.8, 1.0                         \\ \hline
Validation                & \cellcolor[HTML]{EFEFEF}dHCP   & \cellcolor[HTML]{EFEFEF}90 & \cellcolor[HTML]{EFEFEF}27-41                                       & \cellcolor[HTML]{EFEFEF}0.5, 0.8, 1.0 \\ \hline
                          & dHCP                           & 170                        & 30-44                                                               & 0.5, 0.8, 1.0                         \\
\multirow{-2}{*}{Testing} & \cellcolor[HTML]{EFEFEF}M-CRIB & \cellcolor[HTML]{EFEFEF}10 & \cellcolor[HTML]{EFEFEF}40-43                                       & \cellcolor[HTML]{EFEFEF}0.63         
\end{tabular}
\label{tab:datasets}
\end{table}
\vspace{-2ex}
\subsection{Labels}
\vspace{-4ex}
\begin{table}[h!]
\centering
\caption{Subcortical segmentations of the \ac{dHCP}-ALBERTs atlas and matching identifier (ID) equaling the number returned by VINNA for the structure. Left and Right indicate the respective hemisphere. Four  structures are not lateralized (same ID for Left and Right). }
\begin{tabular}{cc|l}
\multicolumn{2}{c|}{\textbf{ID}} & \multicolumn{1}{c}{} \\
\multicolumn{1}{c|}{\textbf{Left}} & \textbf{Right} & \multicolumn{1}{c}{\multirow{-2}{*}{\textbf{Subcortical Structure}}} \\ \hline
\rowcolor[HTML]{EFEFEF} 
\multicolumn{1}{c|}{\cellcolor[HTML]{EFEFEF}0} & 0 & Background \\
\multicolumn{1}{c|}{1} & 2 & Hippocampus \\
\rowcolor[HTML]{EFEFEF} 
\multicolumn{1}{c|}{\cellcolor[HTML]{EFEFEF}3} & 4 & Amygdala \\
\multicolumn{1}{c|}{17} & 18 & Cerebellum \\
\rowcolor[HTML]{EFEFEF} 
\multicolumn{1}{c|}{\cellcolor[HTML]{EFEFEF}19} & 19 & Brainstem \\
\multicolumn{1}{c|}{41} & 40 & Caudate\_nucleus \\
\rowcolor[HTML]{EFEFEF} 
\multicolumn{1}{c|}{\cellcolor[HTML]{EFEFEF}43} & 42 & Thalamus\_high\_intensity\_part\_in\_T2 \\
\multicolumn{1}{c|}{45} & 44 & Subthalamic\_nucleus \\
\rowcolor[HTML]{EFEFEF} 
\multicolumn{1}{c|}{\cellcolor[HTML]{EFEFEF}47} & 46 & Lentiform\_Nucleus \\
\multicolumn{1}{c|}{48} & 48 & Corpus\_Callosum \\
\rowcolor[HTML]{EFEFEF} 
\multicolumn{1}{c|}{\cellcolor[HTML]{EFEFEF}49} & 50 & Lateral-Ventricle \\
\multicolumn{1}{c|}{83} & 83 & CSF \\
\rowcolor[HTML]{EFEFEF} 
\multicolumn{1}{c|}{\cellcolor[HTML]{EFEFEF}84} & 84 & Extra\_cranial\_background \\
\multicolumn{1}{c|}{88} & 85 & IntraCranialBackground \\
\rowcolor[HTML]{EFEFEF} 
\multicolumn{1}{c|}{\cellcolor[HTML]{EFEFEF}87} & 86 & Thalamus\_low\_intensity\_part\_in\_T2
\end{tabular}

\label{tab:labels}
\end{table}

\FloatBarrier
\begin{table*}[h!]
\centering
\caption{Segmentation parcels of the \ac{dHCP}-ALBERTs atlas for gray matter (GM) and white matter (WM). Left and Right indicate the lateralization. The identifier (ID) equals the number returned by VINNA for the structure.}

\begin{tabular}{cc|cc|l}
\multicolumn{2}{c|}{\textbf{GM ID}} & \multicolumn{2}{c|}{\textbf{WM ID}} & \multicolumn{1}{c}{} \\
\multicolumn{1}{c|}{\textbf{Left}} & \textbf{Right} & \multicolumn{1}{c|}{\textbf{Left}} & \textbf{Right} & \multicolumn{1}{c}{\multirow{-2}{*}{\textbf{Structure}}} \\ \hline
\rowcolor[HTML]{EFEFEF} 
\multicolumn{1}{c|}{\cellcolor[HTML]{EFEFEF}5} & 6 & \multicolumn{1}{c|}{\cellcolor[HTML]{EFEFEF}52} & 51 & Anterior\_temporal\_lobe\_medial\_part \\
\multicolumn{1}{c|}{7} & 8 & \multicolumn{1}{c|}{54} & 53 & Anterior\_temporal\_lobe\_lateral\_part \\
\rowcolor[HTML]{EFEFEF} 
\multicolumn{1}{c|}{\cellcolor[HTML]{EFEFEF}9} & 10 & \multicolumn{1}{c|}{\cellcolor[HTML]{EFEFEF}56} & 55 & Gyri\_parahippocampalis\_et\_ambiens\_anterior\_part \\
\multicolumn{1}{c|}{11} & 12 & \multicolumn{1}{c|}{58} & 57 & Superior\_temporal\_gyrus\_middle\_part \\
\rowcolor[HTML]{EFEFEF} 
\multicolumn{1}{c|}{\cellcolor[HTML]{EFEFEF}13} & 14 & \multicolumn{1}{c|}{\cellcolor[HTML]{EFEFEF}60} & 59 & Medial\_and\_inferior\_temporal\_gyri\_anterior\_part \\
\multicolumn{1}{c|}{15} & 16 & \multicolumn{1}{c|}{62} & 61 & Lateral\_occipitotemporal\_gyrus\_gyrus\_fusiformis\_anterior\_part \\
\rowcolor[HTML]{EFEFEF} 
\multicolumn{1}{c|}{\cellcolor[HTML]{EFEFEF}21} & 20 & \multicolumn{1}{c|}{\cellcolor[HTML]{EFEFEF}64} & 63 & Insula \\
\multicolumn{1}{c|}{23} & 22 & \multicolumn{1}{c|}{66} & 65 & Occipital\_lobe \\
\rowcolor[HTML]{EFEFEF} 
\multicolumn{1}{c|}{\cellcolor[HTML]{EFEFEF}25} & 24 & \multicolumn{1}{c|}{\cellcolor[HTML]{EFEFEF}68} & 67 & Gyri\_parahippocampalis\_et\_ambiens\_posterior\_part \\
\multicolumn{1}{c|}{27} & 26 & \multicolumn{1}{c|}{70} & 69 & Lateral\_occipitotemporal\_gyrus\_gyrus\_fusiformis\_posterior\_part \\
\rowcolor[HTML]{EFEFEF} 
\multicolumn{1}{c|}{\cellcolor[HTML]{EFEFEF}29} & 28 & \multicolumn{1}{c|}{\cellcolor[HTML]{EFEFEF}72} & 71 & Medial\_and\_inferior\_temporal\_gyri\_posterior\_part \\
\multicolumn{1}{c|}{31} & 30 & \multicolumn{1}{c|}{74} & 73 & Superior\_temporal\_gyrus\_posterior\_part \\
\rowcolor[HTML]{EFEFEF} 
\multicolumn{1}{c|}{\cellcolor[HTML]{EFEFEF}33} & 32 & \multicolumn{1}{c|}{\cellcolor[HTML]{EFEFEF}76} & 75 & Cingulate\_gyrus\_anterior\_part \\
\multicolumn{1}{c|}{35} & 34 & \multicolumn{1}{c|}{78} & 77 & Cingulate\_gyrus\_posterior\_part \\
\rowcolor[HTML]{EFEFEF} 
\multicolumn{1}{c|}{\cellcolor[HTML]{EFEFEF}37} & 36 & \multicolumn{1}{c|}{\cellcolor[HTML]{EFEFEF}80} & 79 & Frontal\_lobe \\
\multicolumn{1}{c|}{39} & 38 & \multicolumn{1}{c|}{82} & 81 & Parietal\_lobe
\end{tabular}

\label{tab:labels}
\end{table*}
\FloatBarrier


\end{document}